\documentclass[10pt,twocolumn,letterpaper]{article}

\usepackage{cvpr}
\usepackage{times}
\usepackage{epsfig}
\usepackage{graphicx}
\usepackage{amsmath}
\usepackage{amssymb}
\usepackage{longtable}
\usepackage{paralist}
\usepackage{multirow,epstopdf,subfigure}
\usepackage{enumitem}
\usepackage{algorithm}
\usepackage{algorithmic}
\usepackage{bm}

\usepackage[breaklinks=true,bookmarks=false]{hyperref}

\cvprfinalcopy 

\def\cvprPaperID{****} 

\begin{document}

\title{Learning Rate Dropout}

\author{Huangxing Lin\textsuperscript{1}, \ Weihong Zeng\textsuperscript{1}, \  Xinghao Ding\textsuperscript{1*},\ Yue Huang\textsuperscript{1},\ Chenxi Huang \textsuperscript{1}, \ John Paisley\textsuperscript{2}\\
\normalsize \textsuperscript{1}Xiamen University, China\\
\normalsize \textsuperscript{2}Columbia University, New York, NY, USA\\
{\small \tt  $\{hxlin, zengwh \}$@stu.xmu.edu.cn, dxh@xmu.edu.cn,}\\
{\small \tt huangyue05@gmail.com, tongchenhuang@126.com, jpaisley@columbia.edu}
}

\maketitle


\begin{abstract}
	The performance of a deep neural network is highly dependent on its training, and finding better local optimal solutions is the goal of many optimization algorithms. However, existing optimization algorithms show a preference for descent paths that converge slowly and do not seek to avoid bad local optima. In this work, we propose Learning Rate Dropout (LRD), a simple gradient descent technique for training related to coordinate descent. LRD empirically aids the optimizer to actively explore in the parameter space by randomly setting some learning rates to zero; at each iteration, only parameters whose learning rate is not 0 are updated. As the learning rate of different parameters is dropped, the optimizer will sample a new loss descent path for the current update. The uncertainty of the descent path helps the model avoid saddle points and bad local minima. Experiments show that LRD is surprisingly effective in accelerating training while preventing overfitting. 
	(Code:\href{https://github.com/HuangxingLin123/Learning-Rate-Dropout}{https://github.com/HuangxingLin123/Learning-Rate-Dropout}).

\end{abstract}

\section{Introduction}

Deep neural networks are trained by optimizing high-dimensional non-convex loss functions. The success of training hinges on how well we can minimize these loss functions, both in terms of the quality of the convergence points and the time it takes to find them. These loss functions are usually optimized using gradient-descent-based algorithms. These optimization algorithms search the descending path of the loss function in the parameter space according to a predefined paradigm. Existing optimization algorithms can be roughly categorized into adaptive and non-adaptive methods. Specifically, adaptive algorithms (Adam \cite{Kingma2014Adam}, Amsgrad \cite{reddi2019convergence}, RMSprop \cite{tieleman2012lecture}, RAdam \cite{liu2019variance}, \emph{etc}) tend to find paths that descend the loss quickly, but usually converge to bad local minima. In contrast, non-adaptive algorithms (\emph{e.g.} SGD-momentum) converge better, but the training speed is slow. Helping the optimizer maintain fast training speed and good convergence is an open problem.

\begin{figure}
	\centering
	
	\subfigure[]{\includegraphics[width=1.5in]{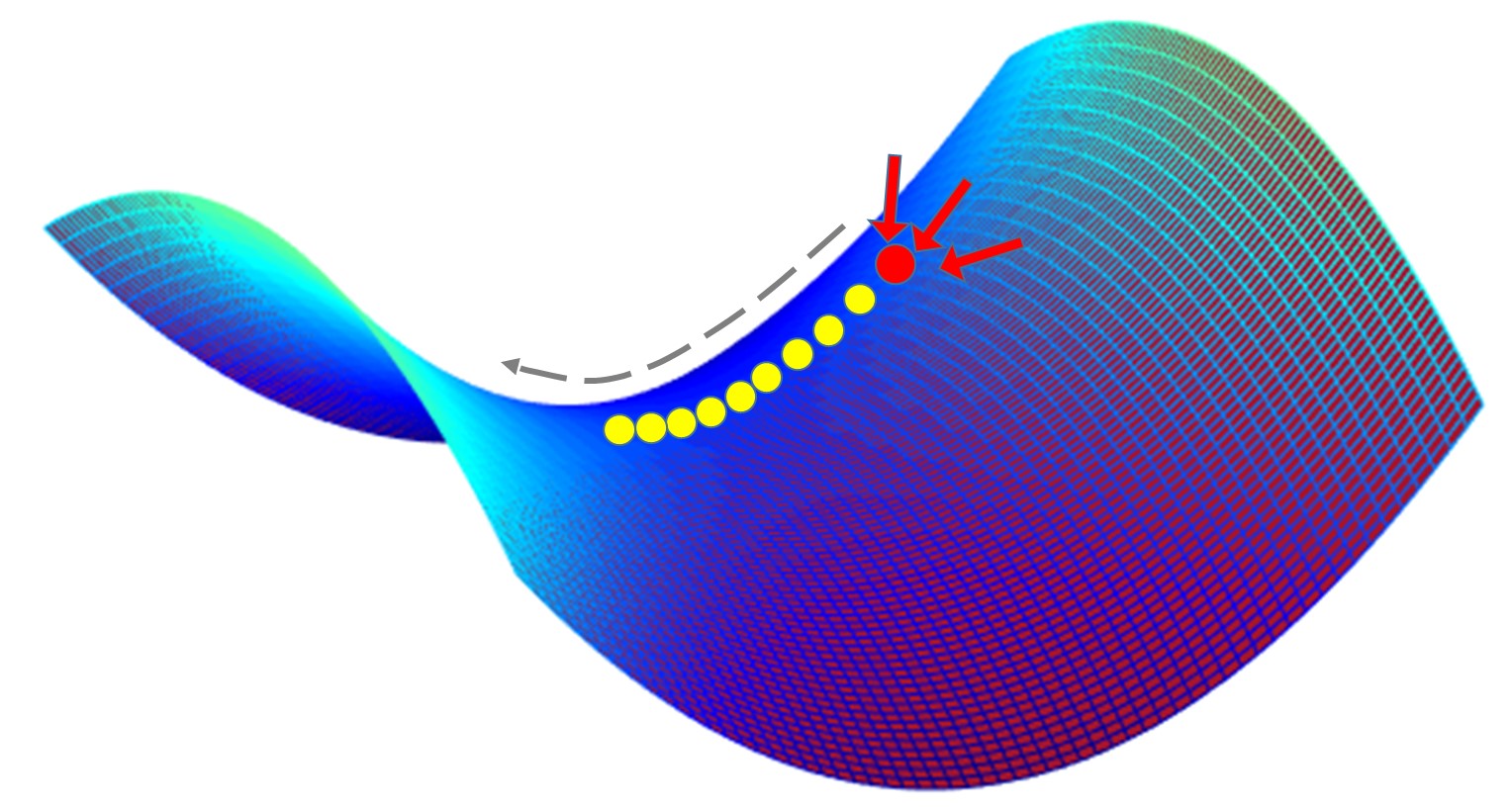}}
	\subfigure[]{\includegraphics[width=1.5in]{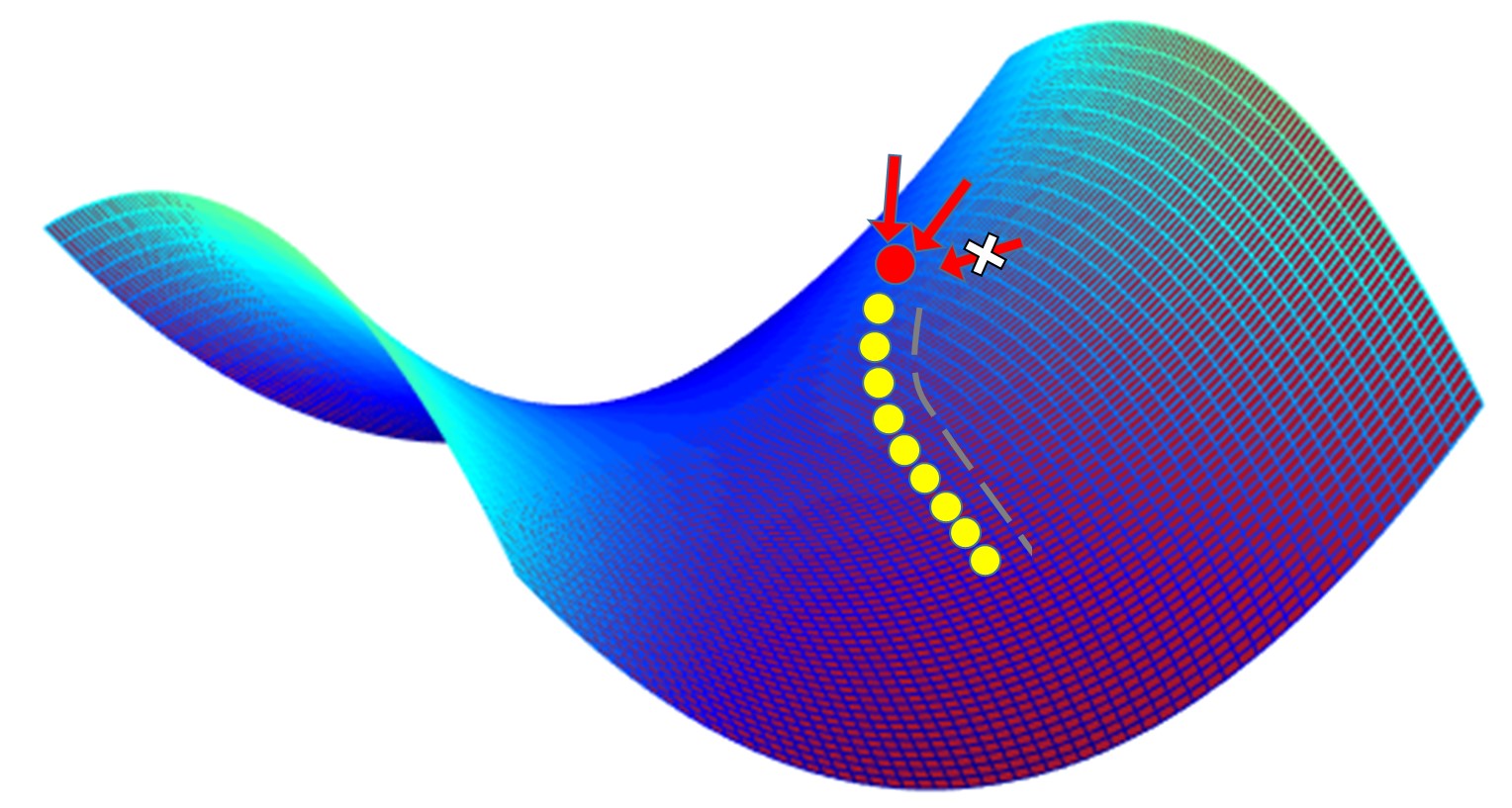}}
	\caption{(a) Gradient updates are trapped in a saddle point. (b) Applying learning rate dropout to training, the optimizer escapes from saddle points more quickly. Red arrow: the update of each parameter in current iteration. Red dot: the initial state. Yellow dots: the subsequent states. $\bm{\times}$: randomly dropped learning rate.}
	\label{fig1}
\end{figure}

\begin{figure*}[t]
	\centering
	\subfigure[Back propagation]{\includegraphics[width=1.6in]{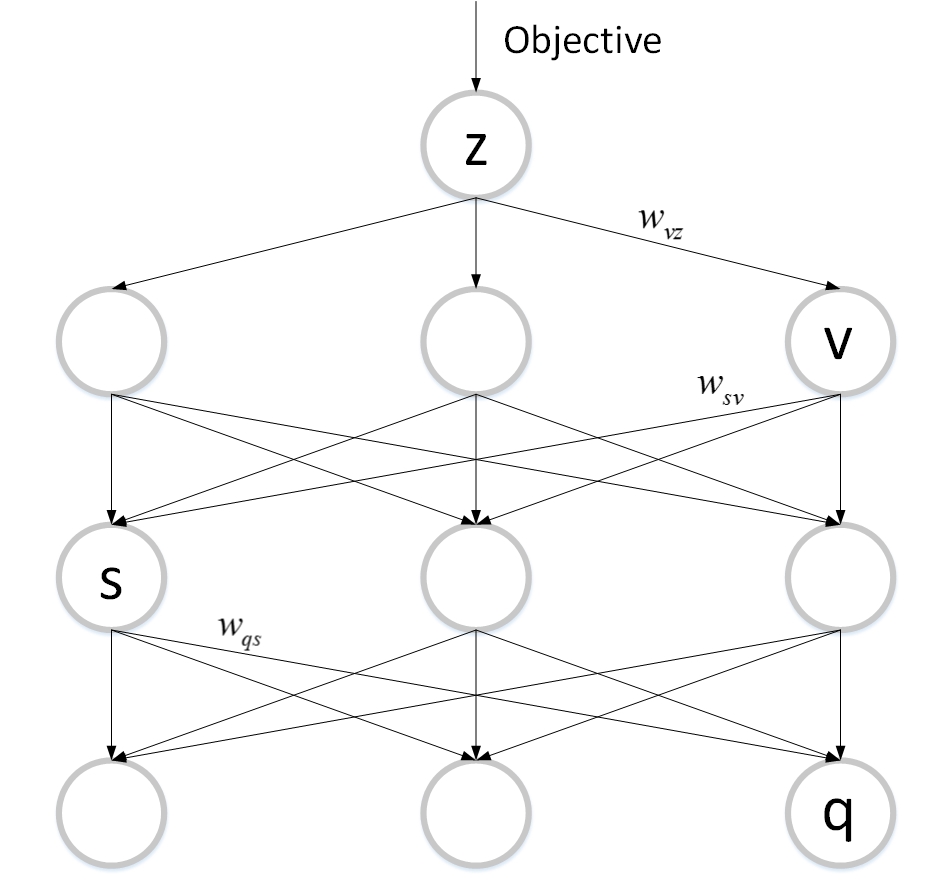}}
	\hspace{1cm}
	\subfigure[Applying dropout]{\includegraphics[width=1.6in]{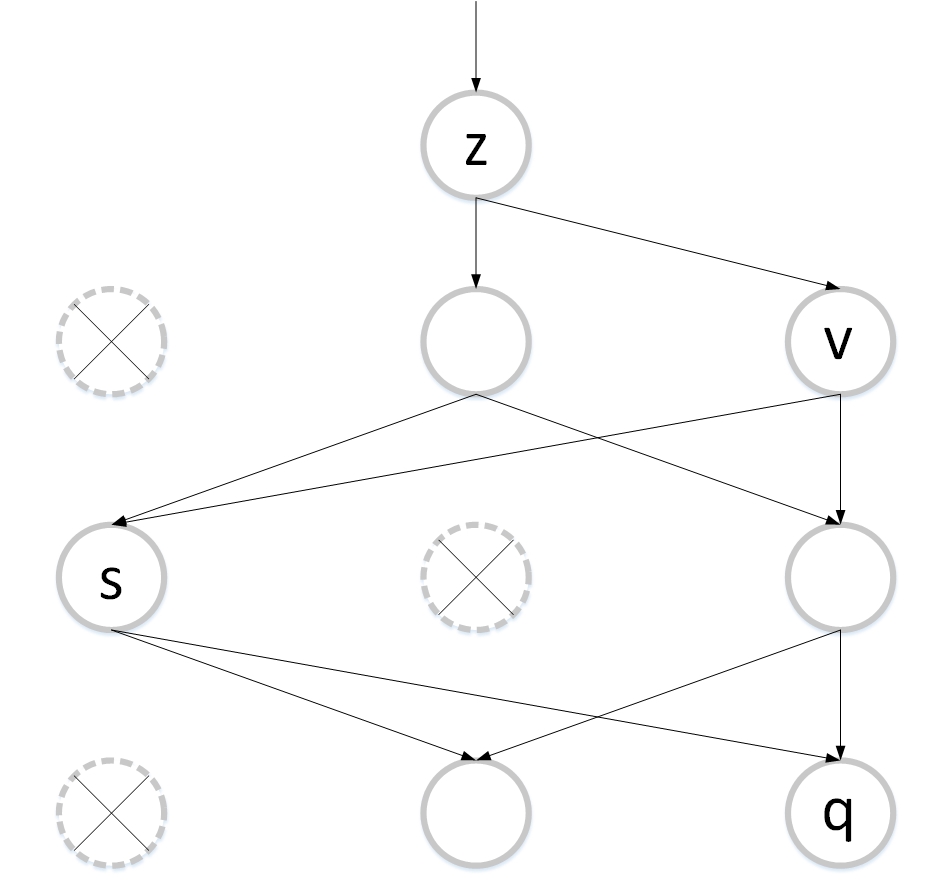}}
	\hspace{1cm}
	\subfigure[Applying learning rate dropout]{\includegraphics[width=1.6in]{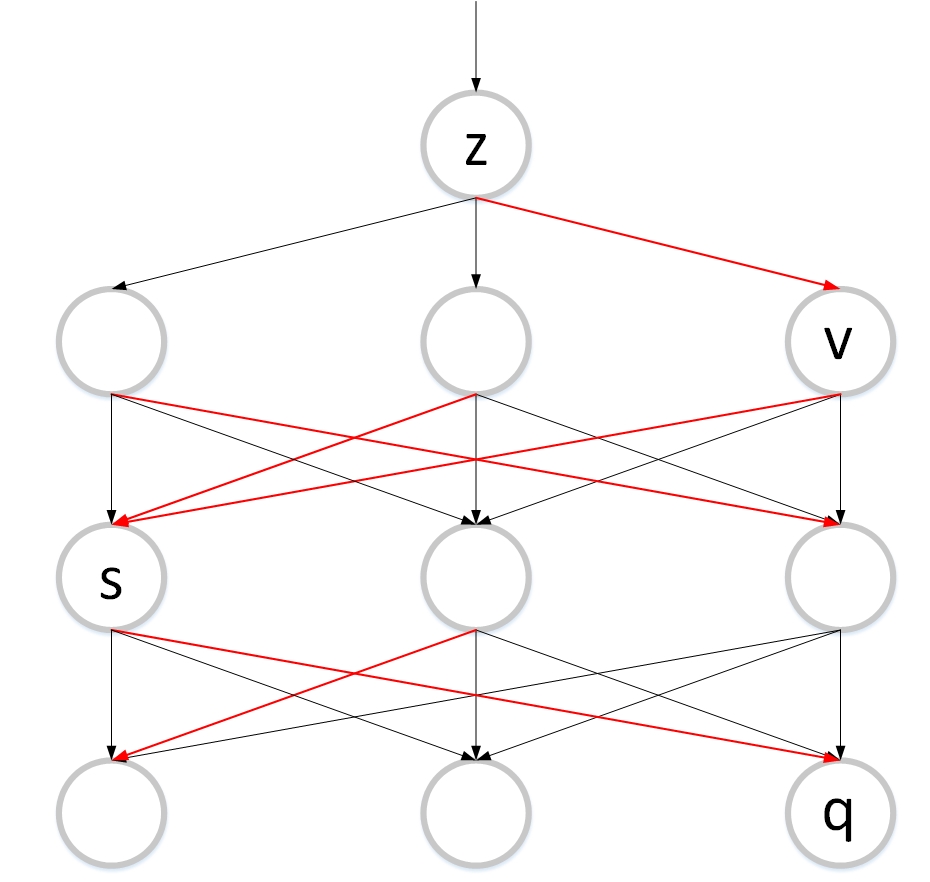}}
	
	\caption{(a) BP algorithm for a neural network. Black lines represent the gradient updates to each weight parameter (\emph{e.g.} $w_{qs}$, $w_{sv}$, $w_{vz}$). (b) An example of applying standard dropout, the dropped units do not appear in both forward and back propagation during training. (c) The red line indicates that the learning rate is dropped, so the corresponding weight parameter is not updated. Note that the dropped learning rate does not affect forward propagation and gradient back propagation. At each iteration, different learning rates are dropped.}
	\label{bp}
\end{figure*}

Implicit regularization techniques (\emph{e.g.} dropout \cite{srivastava2014dropout}, weight decay \cite{krizhevsky2009learning}, noisy label \cite{xie2016disturblabel}) are widely used to help training. The most popular one is dropout, which can prevent feature co-adaptation (a sign of overfitting) effectively by randomly dropping the hidden units (\emph{i.e.} their activation is zeroed). Dropout can be interpreted as a way of regularizing training by adding noise to the hidden units. Other methods can achieve similar effects by injecting noise into gradients \cite{neelakantan2015adding}, label \cite{xie2016disturblabel} and activation functions \cite{gulcehre2016noisy}. The principle of these methods is to force the optimizer to randomly change the loss descent path in the high-dimensional parameter space by injecting interference into the training. The uncertainty of the loss descent path gives the model more opportunities to escape local minima and find a better result. However, many researchers have found that these 
regularization methods can improve generalization at the cost of training time \cite{srivastava2014dropout, wang2013fast}. This is because random noise in training may create an erroneous loss descent path (\emph{i.e.} current parameter updates may increase the loss), resulting in slow convergence. In fact, what is needed is a technique that finds good local optima and converges quickly. 

A possible solution is to inject uncertainty into the loss descent path while ensuring that the resulting path is correct. A correct loss descent path means it decreases the loss. On the other hand, inspired by the coordinate descent algorithm \cite{wright2015coordinate, nesterov2012efficiency, necoara2013random}, we realize a fundamental fact that the decline of loss does not depend solely on the simultaneous update of all weight parameters. According to coordinate descent, even if only one parameter is updated, the loss is reduced. In particular, the loss descent path is determined by all updated parameters. This means that we can sample a new descent path by randomly pausing the update of certain parameters. Based on these observations, we propose a new regularization technique, \emph{learning rate dropout} (LRD), to help the optimizer randomly sample the correct loss descent path at each iteration.

Learning rate dropout provides a way to inject randomness into the loss descent path while maintaining a descent direction. The key difference from standard dropout is to randomly drop the learning rate of the model parameters instead of the hidden units. During training, the optimizer dynamically calculates the update and allocates a learning rate for each parameter. Learning rate dropout works by randomly determining which parameters are not updated (\emph{i.e.} the learning rate is set to zero) for the current training iteration. In the simplest case, the learning rate for each parameter is retained with a fixed probability $p$, independently from other parameters, or is dropped with probability $1-p$. A dropped learning rate is temporarily set to zero, which does not affect the update of other parameters or subsequent updates. 
A neural network with $n$ parameters has $2^n$ feasible descent paths. Since each path is correct in that it decreases the objective function, our learning rate dropout does not hinder training unlike previous methods \cite{srivastava2014dropout, xie2016disturblabel}. However, convergence issues such as saddle points and bad local optima are avoided more easily using the proposed method. Furthermore, LRD can be trivially integrated into existing gradient-based optimization algorithm, such as Adam.

\begin{algorithm*}[t] 
	\caption{Generic framework of optimization with learning rate dropout. $\circ$ indicates element-wise multiplication.} 
	\label{algorithm1} 
	\begin{algorithmic}[1] 
		\REQUIRE 
		$\alpha:$ learning rate, $\left\{\phi_t,\psi_t\right\}_{t=1}^T$$:$ function to calculate momentum and adaptive rate, $W_0:$ initial parameters, \\ $f(W):$ stochastic objective function, $p:$ dropout rate, $AdaptiveMethod:$ False or True.
		\ENSURE $W_T:$ resulting parameters.
		
		\FOR{$t=1$ \textbf{to} $T$} 
		
		\STATE $G_t=\triangledown f_t(W_{t-1})$   (Calculate gradients w.r.t. stochastic objective at timestep $t$)
		
		\STATE $M_t={\phi}_t(G_1,\cdot \cdot \cdot,G_t)$ (Accumulation of past and current gradients)
		
		\IF{$AdaptiveMethod$ is \textbf{\emph{True}}} 
		\STATE $V_t=\psi_t(G_1,\cdot \cdot \cdot,G_t)$  (Accumulation of squared gradients)
		\STATE $\triangle W_t=M_t/\sqrt{V_t}$ 
		\ELSE 
		\STATE $\triangle W_t=M_t$ 
		
		\ENDIF
		
		\STATE	Random sample learning rate dropout mask $D_t$ with each element $d_{ij,t} \sim Bernoulli(p)$ 
		\STATE $A_t=\alpha D_t$ (Randomly drop learning rates at timestep $t$)
		\STATE $U_t=A_t \circ \triangle W_t$ (Calculate the update for each parameter)
		\STATE $W_t=W_{t-1}-U_t$
		
		\ENDFOR
		
	\end{algorithmic}
\end{algorithm*}

\section{Related work}
\paragraph{Noise injection:} Learning rate dropout can be interpreted as a regularization technique for loss descent path by adding noise to the learning rate. The idea of adding noise \cite{adilova2018introducing,hinton2003stochastic,graves2011practical,vincent2010stacked} to the training of neural networks has drawn much attention. Adilova, \emph{et al}. \cite{adilova2018introducing} demonstrate that noise injection substantially improves model quality for non-linear neural network. An \cite{an1996effects} explored the effects of noise injection to the inputs, outputs, and weights of multilayer feedforward neural networks. Blundell, \emph{et al}. \cite{blundell2015weight} regularize the weights by minimizing a variational free energy. Neelakantan, \emph{et al}. \cite{neelakantan2015adding} also find that adding noise to gradients can help avoid overfitting and result in lower training loss.
Xie, \emph{et al}. \cite{xie2016disturblabel} imposes regularization within the loss layer by randomly setting the labels to be incorrect. Similarly, the standard dropout \cite{srivastava2014dropout} is a way of regularizing a neural network by adding noise to its hidden units. Wan, \emph{et al}. \cite{wan2013regularization} further proposed DropConnect, which is a generalization of Dropout. DropConnect sets a randomly selected subset of weights to zero, rather than hidden units. On the other hand, adding noise to activation functions can prevent the early saturation. Gulcehre, \emph{et al}. \cite{gulcehre2016noisy} found that noisy activation functions (\emph{e.g.}, sigmoid and tanh) are easier to optimize. Replacing the non-linearities by their noisy counterparts usually leads to better results. Chen, \emph{et al}. \cite{chen2017noisy} propose a noisy softmax to mitigate the early saturation issue by injecting annealed noise to the softmax input. 
All of these methods can effectively prevent overfitting. However, they result in slower convergence.

\paragraph{Optimization algorithms:} Deep neural networks are optimized using gradient-descent-based algorithms. Stochastic gradient descent (SGD) \cite{robbins1951stochastic} is a widely used approach, which performs well in many research fields. However, it has been empirically observed that SGD has slow convergence since it scales the gradient uniformly in all directions. To address this issue, variants of SGD that adaptively rescale or average the gradient direction have achieved some success. Examples include RMSprop \cite{tieleman2012lecture}, Adadelta \cite{zeiler2012adadelta} and Adam \cite{Kingma2014Adam}. In particular, Adam is the most popular adaptive optimization algorithm due to its rapid training speed. However, many publications \cite{wilson2017marginal, reddi2019convergence} indicate that Adam has poor convergence. Recently, Amsgrad \cite{reddi2019convergence} and Adabound \cite{luo2019adaptive}, two variants of Adam, were proposed to solve the convergence issues of Adam by bounding the learning rates. Furthermore, RAdam \cite{liu2019variance} also aims to solve the convergence issue of Adam by rectifying the variance of the adaptive learning rate. While these adaptive methods often display faster progress in training, they have also been observed to fail to converge well in many cases. 

\section{Method description}
\subsection{Online optimization}
We start our discussion on learning rate dropout by integrating it into an online optimization problem \cite{zinkevich2003online}. We provide a generic framework of optimization methods with learning rate dropout in Algorithm \ref{algorithm1} (all multiplications are element-wise). Consider a neural network with weight parameters $W$. We assume $W\in \mathbb{R}^{h\times k}$. In the online setup, at each time step $t$, the optimization algorithm modifies the current parameters $W_{t-1}$ using the loss function over data time $t$, $f_t(W_{t-1})$ and gradient $G_t=\triangledown f_t(W_{t-1})$. Finally, the optimizer calculates the update for $W_{t-1}$ using $G_t$ and possibly other terms including earlier gradients.


Algorithm \ref{algorithm1} encapsulates many popular adaptive and non-adaptive methods by the definition of gradient accumulation terms $\phi(\cdot)$ and $\psi(\cdot)$. In particular, the adaptive methods are distinguished by the choice of $\psi(\cdot)$, which is absent in non-adaptive methods. Most methods contain the similar momentum component
\begin{equation}
\begin{aligned}
M_t & =\phi_t(G_1,\cdot \cdot \cdot,G_t)\\
&=\beta M_{t-1}+\eta G_t,
\end{aligned}
\label{eq.1}
\end{equation}
where $\beta>0$, $\eta>0$, $\beta$ is the momentum parameter. The momentum accumulates the exponentially moving average of previous gradients to correct the current update.


\subsection{Learning rate dropout}
During training, the optimizer assigns a learning rate $\alpha$, which is typically a constant, to each parameter. Applying learning rate dropout in optimization, in one iteration the learning rate of each parameter is kept with probability $p$, independently from other parameters, or set to zero otherwise. At each time step $t$, a random binary mask matrix $D_t \in \{0,1\}^{h\times k}$ is sampled to encode the learning rate information with each element $d_{ij,t} \sim Bernoulli(p)$. Then a learning rate matrix $A_t$ at time step $t$ is obtained by 
\begin{equation}
\begin{aligned}
A_t=\alpha D_t.\\
\end{aligned}
\end{equation} 

\begin{figure}
	\centering
	
	\includegraphics[width=3in]{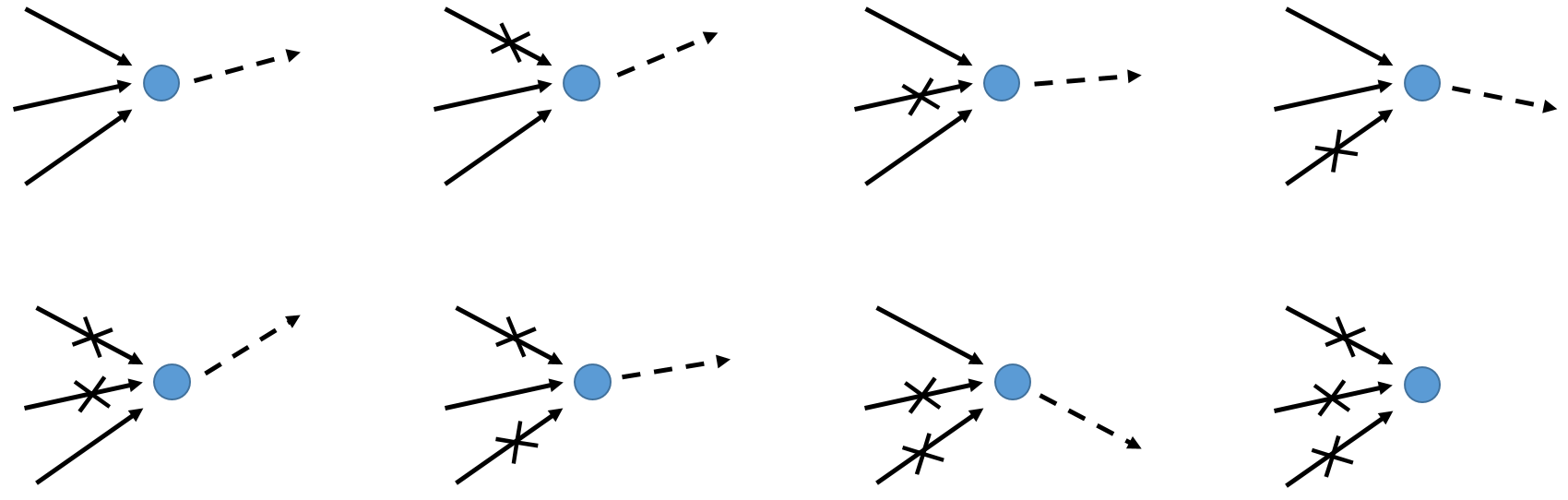}
	
	\caption{Applying learning rate dropout to training. This model contains 3 parameters, so there are 8 loss descent paths to choose in each iteration. The blue dot is the model state. The solid line is the update of each parameter. The dashed line is the resulting update for the model. ``$\bm{\times}$'' represents dropping the learning rate.}
	\label{fig2}
\end{figure}

\paragraph{LRD and coordinate descent:} For a parameter, a learning rate of $0$ means not updating its value. For a model, applying learning rate dropout is equivalent to uniformly sampling one of $2^n$ possible parameter subsets for which to perform a gradient update in the current iteration (see Figure \ref{fig2}). This is closely connected to coordinate descent, in which each set of a partition of parameters is fully optimized in a cycle, before returning to the first set for the next iteration. Therefore, each update of LRD will still cause a decrease in the loss function according to the gradient descent theory \cite{rumelhart1985learning, wright2015coordinate, nesterov2012efficiency}, while better avoiding saddle points. On the other hand, our LRD does not interrupt the gradient calculation of any parameters. If there is a gradient accumulation term (\emph{e.g.} momentum) in the optimizer, the gradients of each parameter will be stored for subsequent updates, regardless of whether the learning rate is dropped. Furthermore, our method won't slow training like previous methods \cite{srivastava2014dropout,xie2016disturblabel,wan2013regularization}.

\begin{figure}[t]
	\centering
	
	\subfigure[Adam]{\includegraphics[width=1.6in]{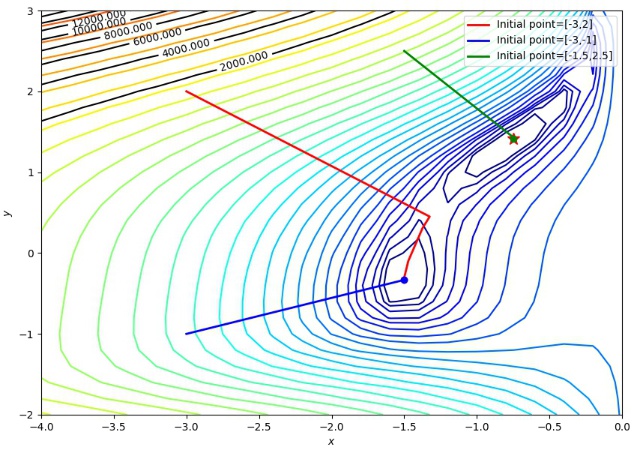}}
	\subfigure[Adam with learning rate dropout]{\includegraphics[width=1.6in]{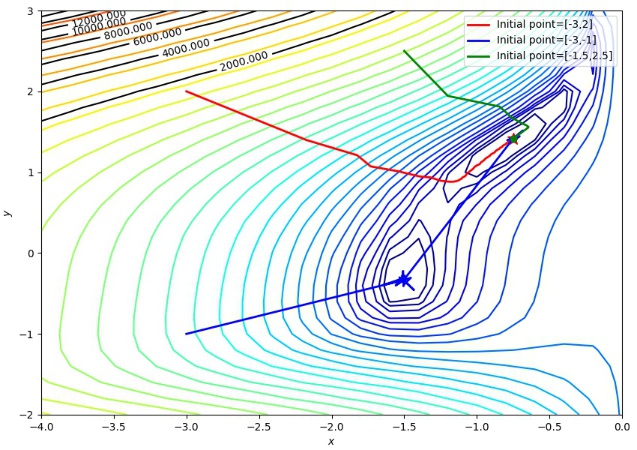}}
	\caption{Visualization of the loss descent paths. The learning rate dropout can help Adam escape from the local minimum. ``$\star$'' is optimal point $(-0.74,1.40)$.}
	\label{fig3}
\end{figure}

\paragraph{Toy example:} The learning rate dropout can accelerate training and improve generalization. By adding stochasticity to loss descent path, this technique helps the model to traverse quickly through the ``transient'' plateau (\emph{e.g.} saddle points or local minima) and gives the model more chances to find a better minimum. To verify the effectiveness of learning rate dropout, we show a toy example to visualize the loss descent path during optimization. Consider the following nonconvex function:
\begin{equation}
\begin{aligned}
f(x,y)=&(1.5-x^2+xy)^2+(2.25-x^2+xy^2)^2\\
&+(2.625-x^2+xy^3)^2,\\
\end{aligned}
\end{equation} 
where $x\in[-4,0]$, $y\in[-2.0,3.0]$.  We use the popular optimizer Adam to search the minima of the function in the two dimensional parameter space. For this function, the point $(-0.74,1.40)$ is the optimal solution. In Figure \ref{fig3}, we see that the convergence of Adam is sensitive to the initial point. Different initializations lead to different convergence results. And, Adam is easily trapped by the minimum near the initial point, even if it is a bad local minimum. In contrast, our learning rate dropout makes Adam more active. Even if the optimizer reach a local minimum, the learning rate dropout still encourages the optimizer to search for other possible paths instead of doing nothing. This example illustrates that the learning rate dropout can effectively help the optimizer escape from suboptimal points and find a better result.

\begin{figure*}
	\centering
	\includegraphics[width=.19\textwidth]{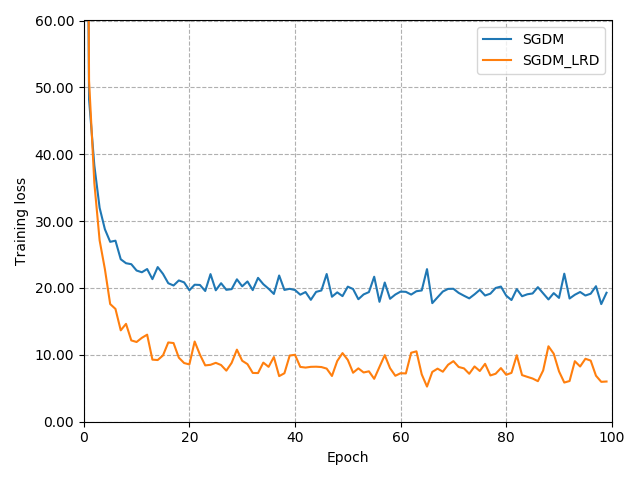}
	\includegraphics[width=.19\textwidth]{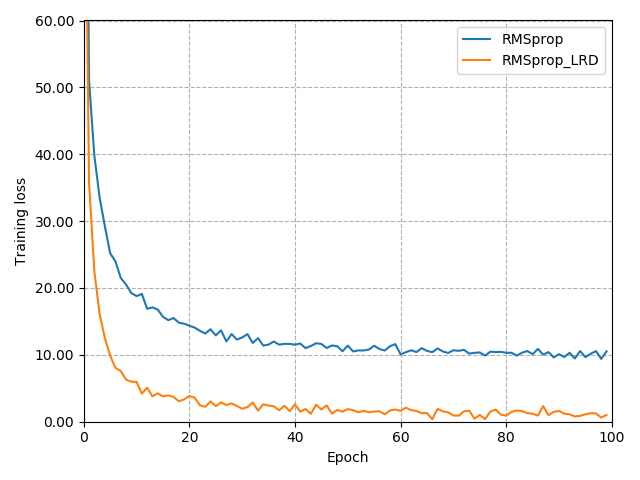}
	\includegraphics[width=.19\textwidth]{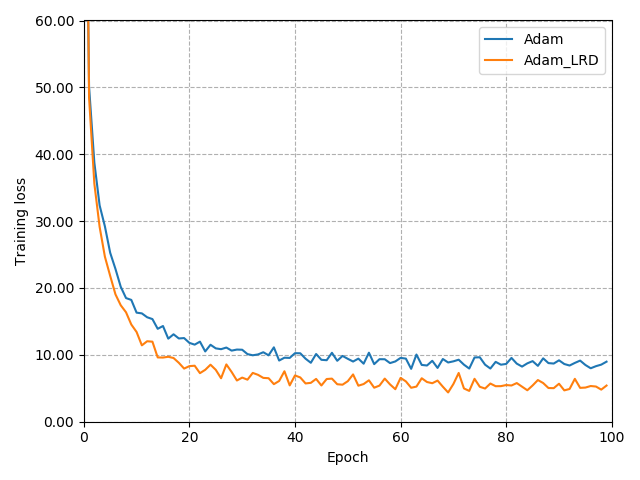}
	\includegraphics[width=.19\textwidth]{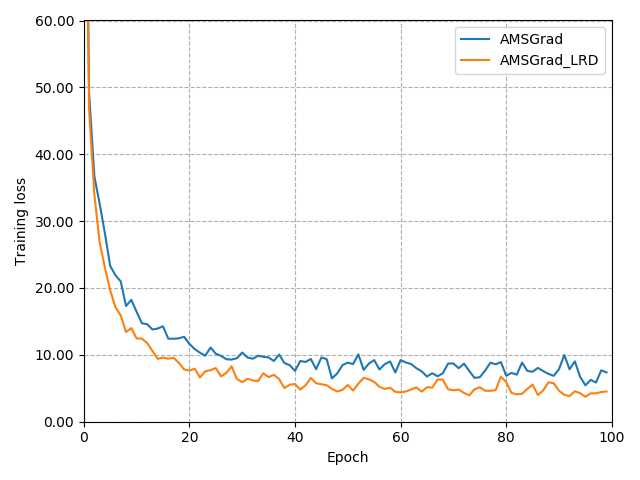}
	\includegraphics[width=.19\textwidth]{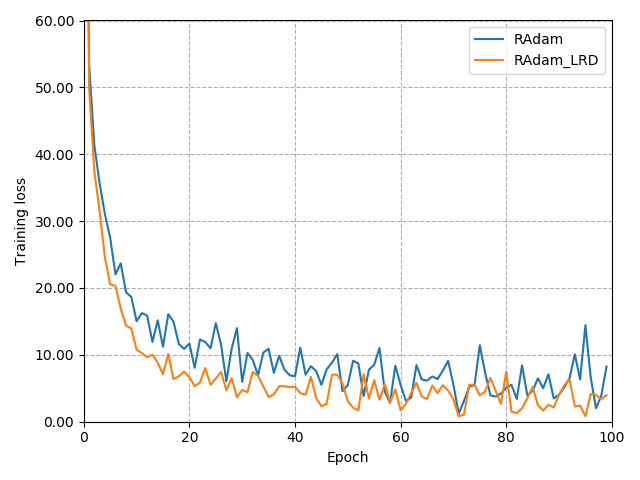}\\  
	\includegraphics[width=.19\textwidth]{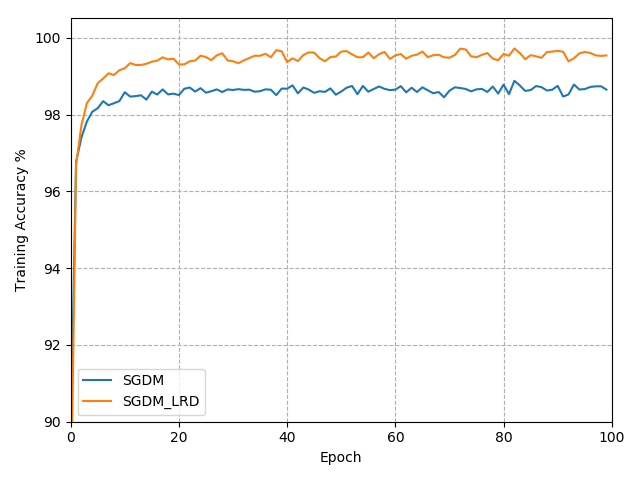}
	\includegraphics[width=.19\textwidth]{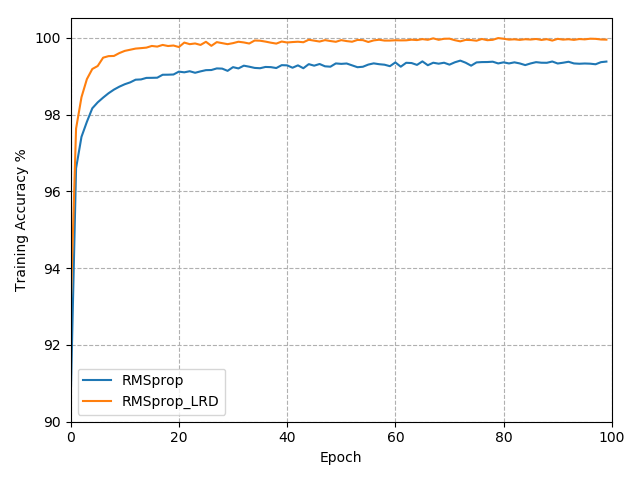}
	\includegraphics[width=.19\textwidth]{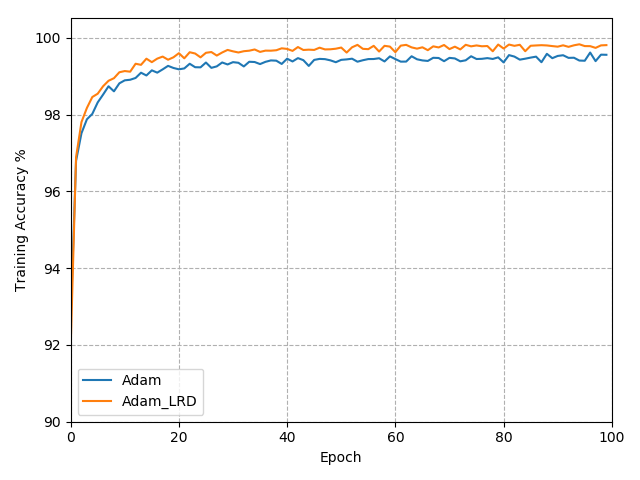}
	\includegraphics[width=.19\textwidth]{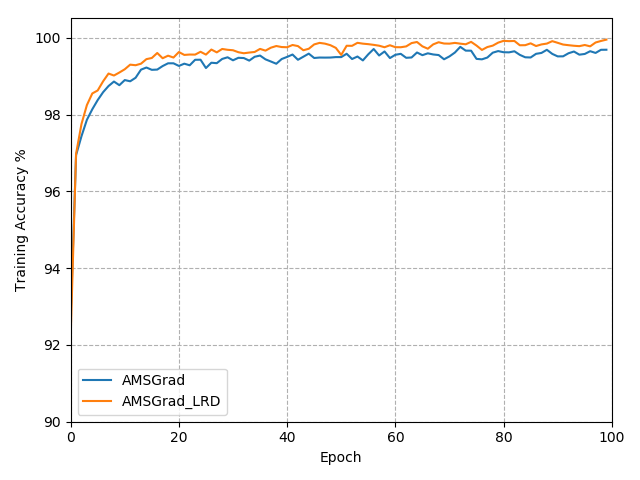}
	\includegraphics[width=.19\textwidth]{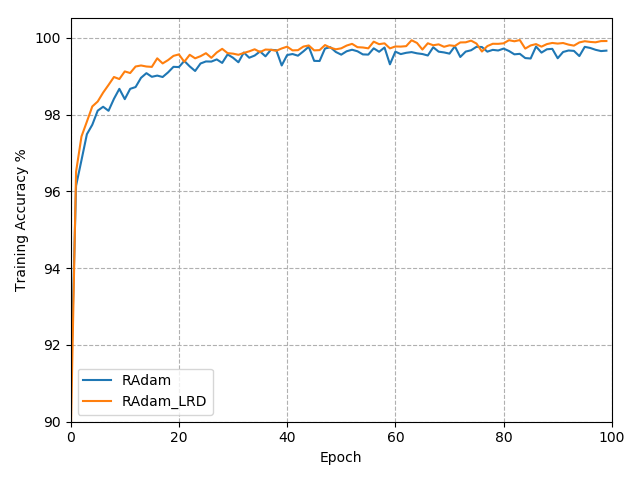}\\  
	\includegraphics[width=.19\textwidth]{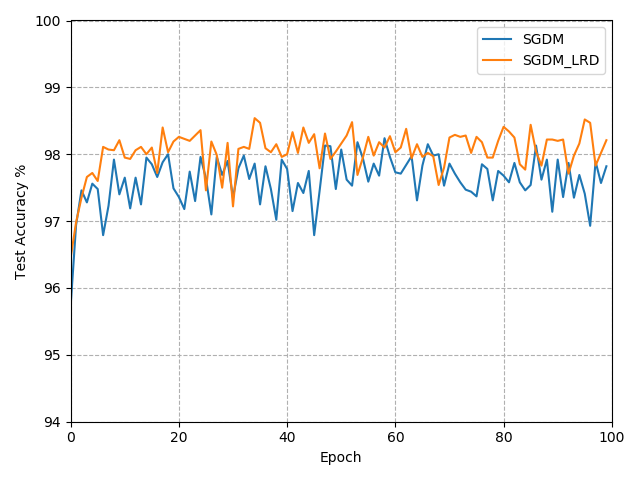}
	\includegraphics[width=.19\textwidth]{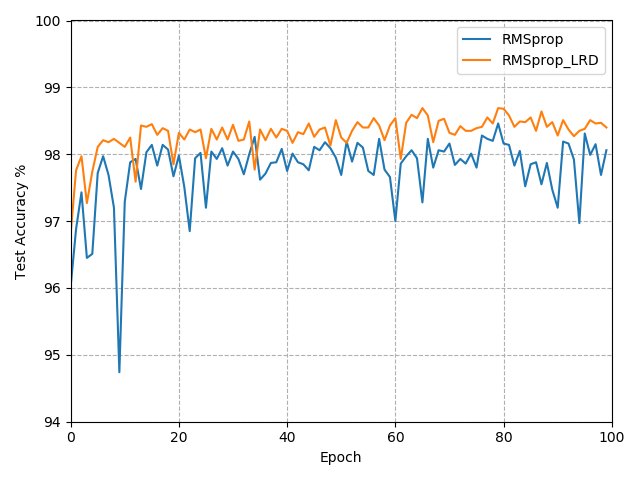}
	\includegraphics[width=.19\textwidth]{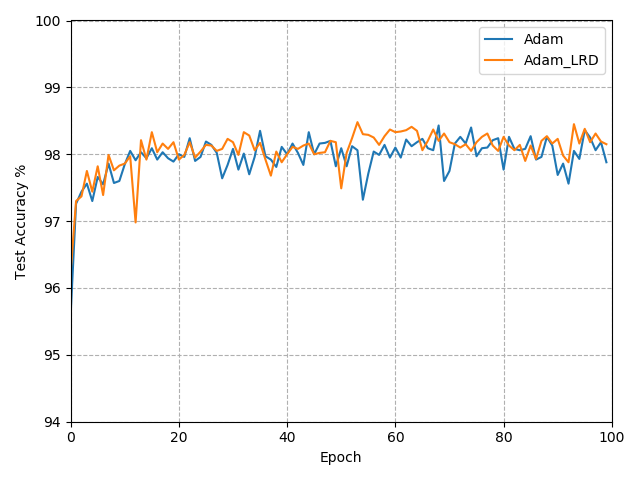}
	\includegraphics[width=.19\textwidth]{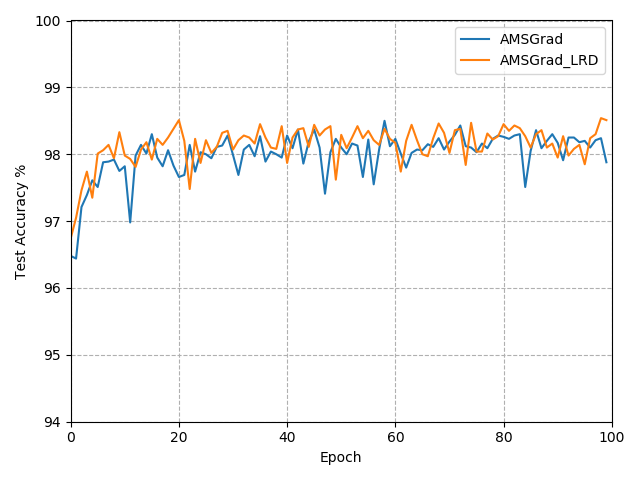}
	\includegraphics[width=.19\textwidth]{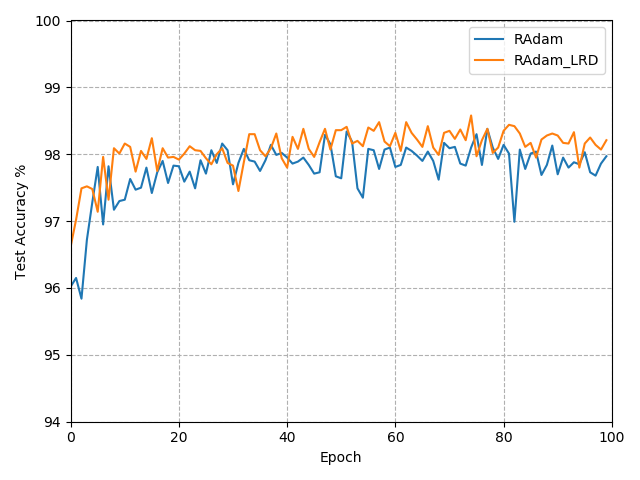}
	\caption{The learning curves for 2-layers FCNet on MNIST. Top: Training loss. Middle:  Training accuracy. Bottom: Test accuracy.}
	\label{mnist}
\end{figure*}

\begin{figure*}
	\centering
	\includegraphics[width=.19\textwidth]{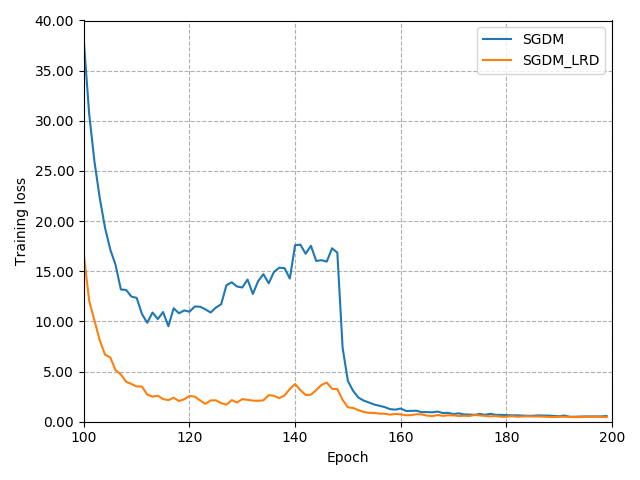}
	\includegraphics[width=.19\textwidth]{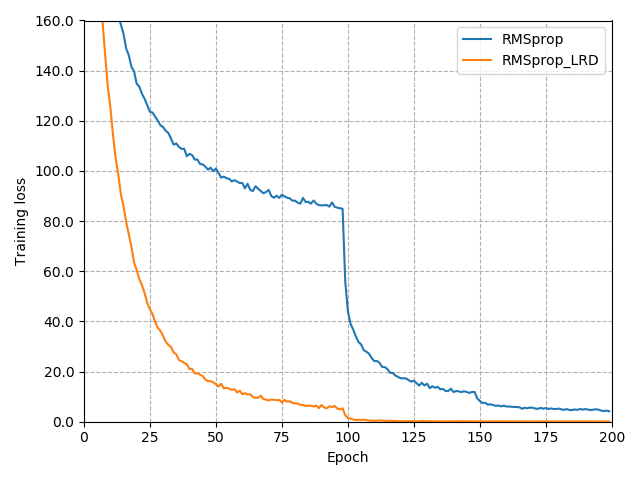}
	\includegraphics[width=.19\textwidth]{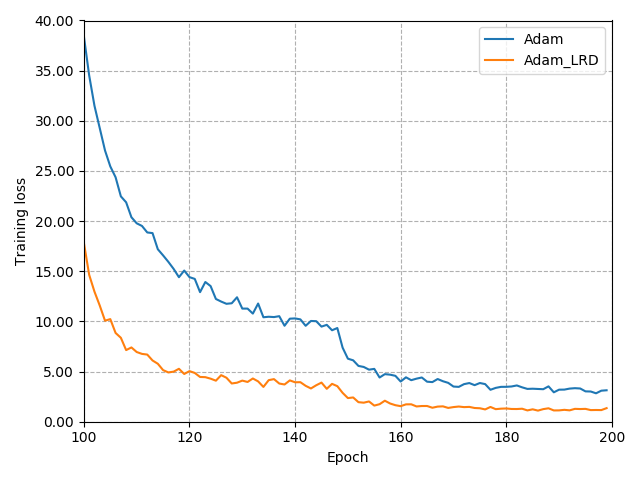}
	\includegraphics[width=.19\textwidth]{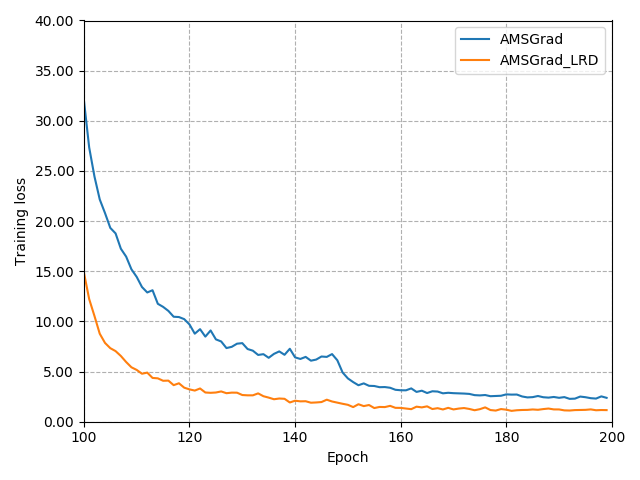}
	\includegraphics[width=.19\textwidth]{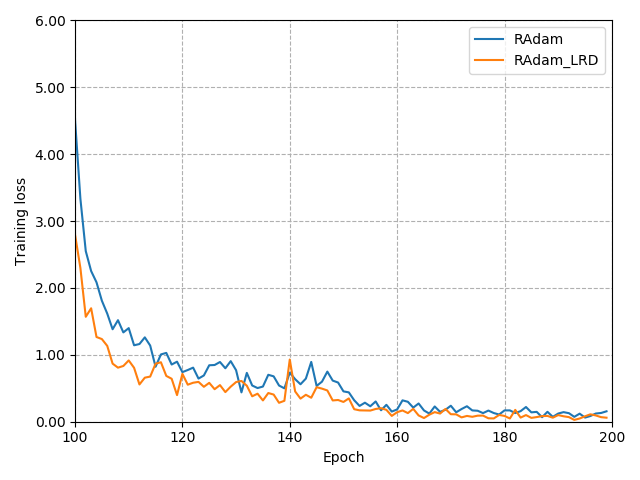}\\  
	\includegraphics[width=.19\textwidth]{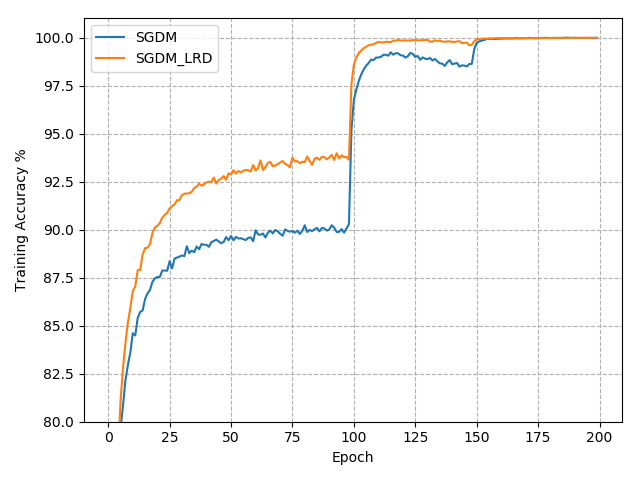}
	\includegraphics[width=.19\textwidth]{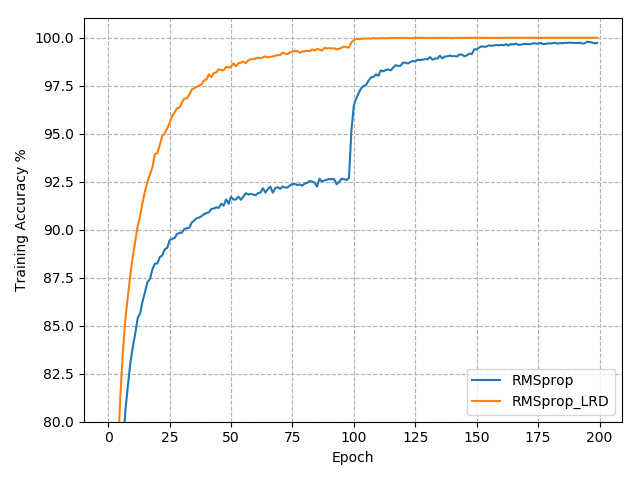}
	\includegraphics[width=.19\textwidth]{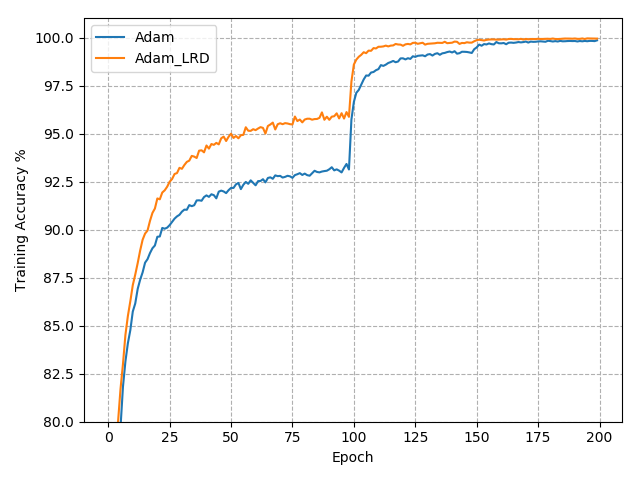}
	\includegraphics[width=.19\textwidth]{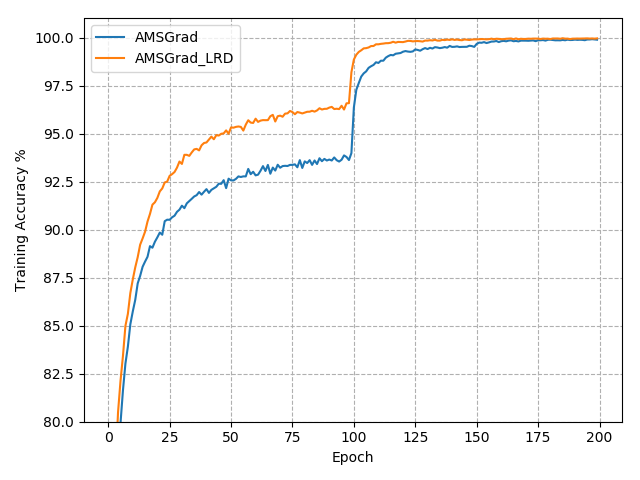}
	\includegraphics[width=.19\textwidth]{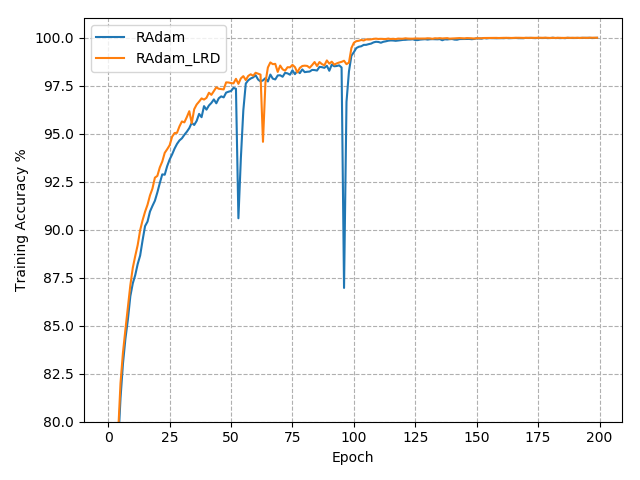}\\  
	\includegraphics[width=.19\textwidth]{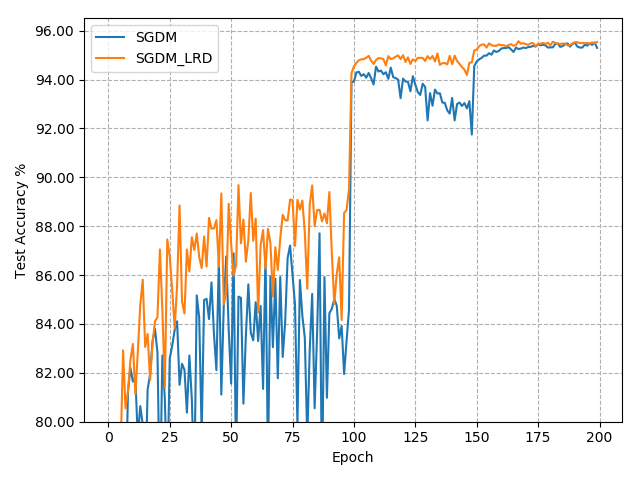}
	\includegraphics[width=.19\textwidth]{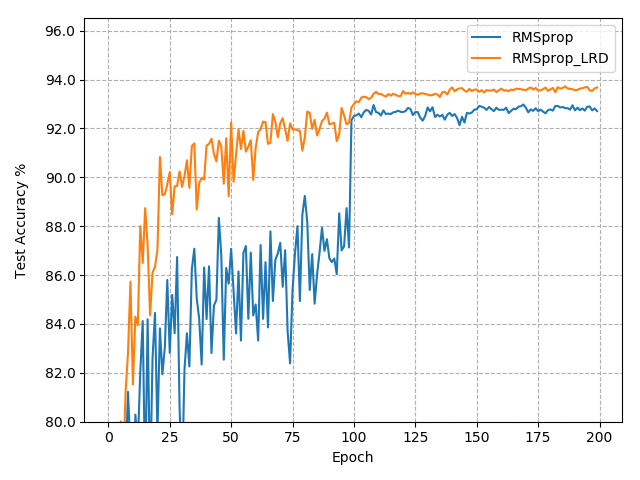}
	\includegraphics[width=.19\textwidth]{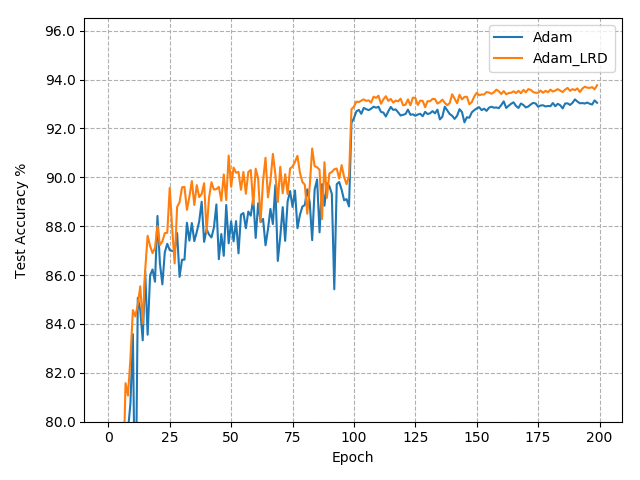}
	\includegraphics[width=.19\textwidth]{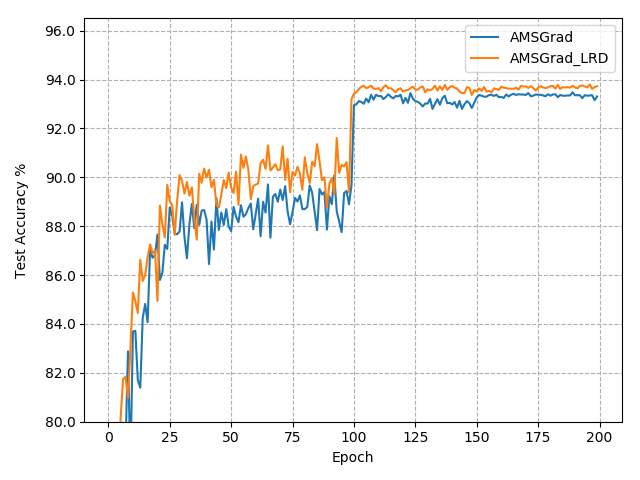}
	\includegraphics[width=.19\textwidth]{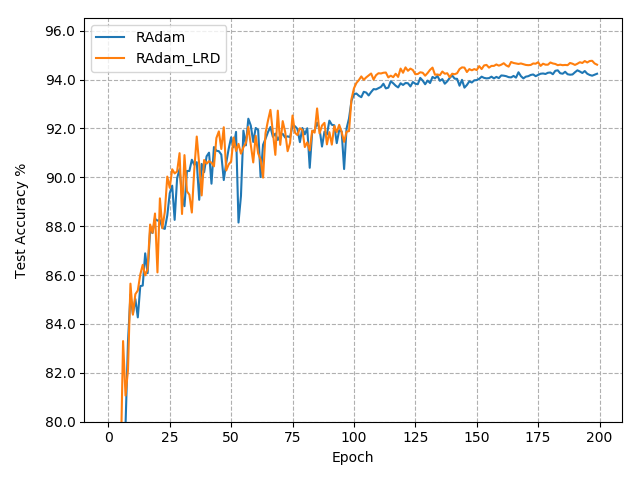}
	\caption{The learning curves for ResNet-34 on CIFAR-10. Top: Training loss. Middle:  Training accuracy. Bottom: Test accuracy.}
	\label{cifar10}
\end{figure*}

\section{Experiments}
To empirically evaluate the proposed method, we perform different popular machine learning tasks, including image classification, image segmentation and object detection. Using different models and optimization algorithms, we demonstrate that learning rate dropout is a general technique for improving neural network training not specific to any particular application domain. 

\begin{figure*}
	\centering
	\includegraphics[width=.19\textwidth]{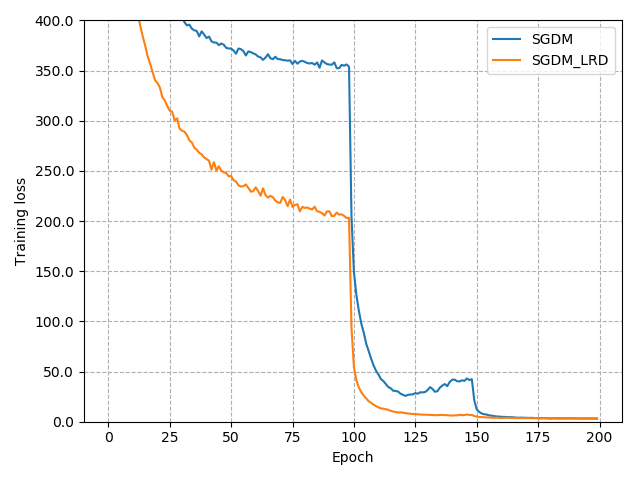}
	\includegraphics[width=.19\textwidth]{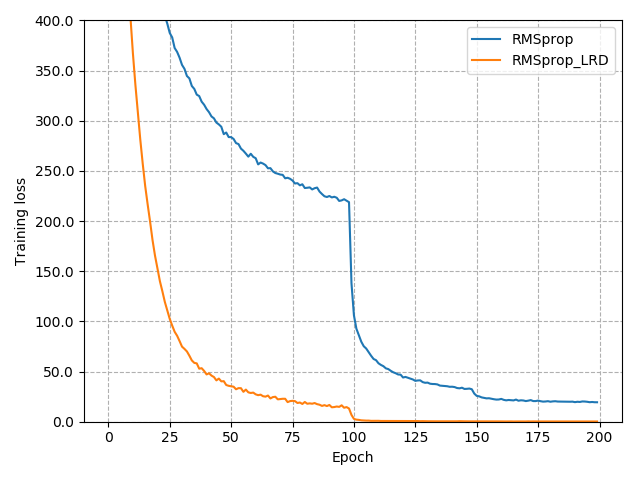}
	\includegraphics[width=.19\textwidth]{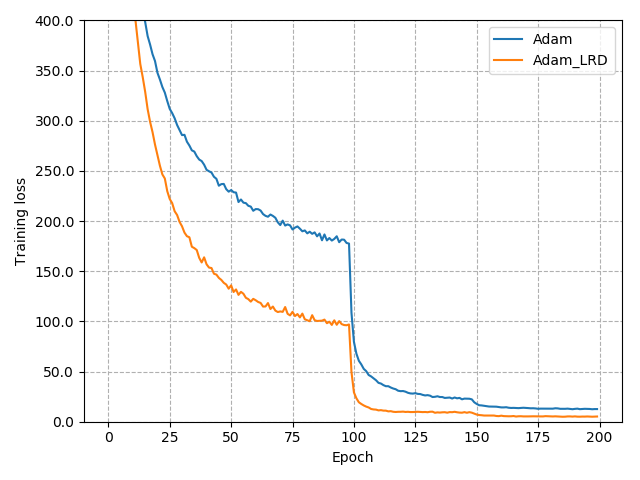}
	\includegraphics[width=.19\textwidth]{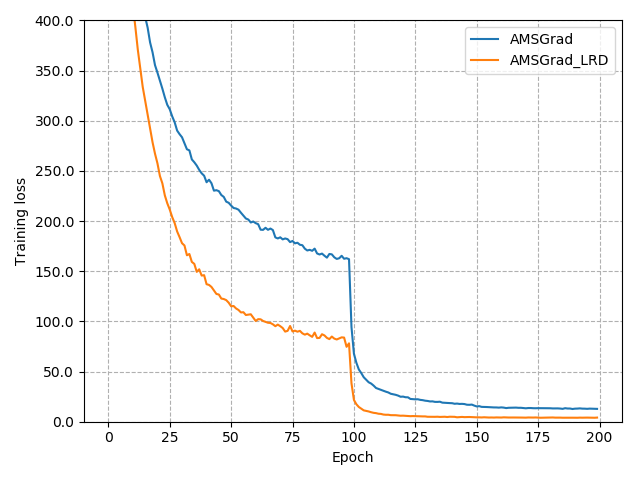}
	\includegraphics[width=.19\textwidth]{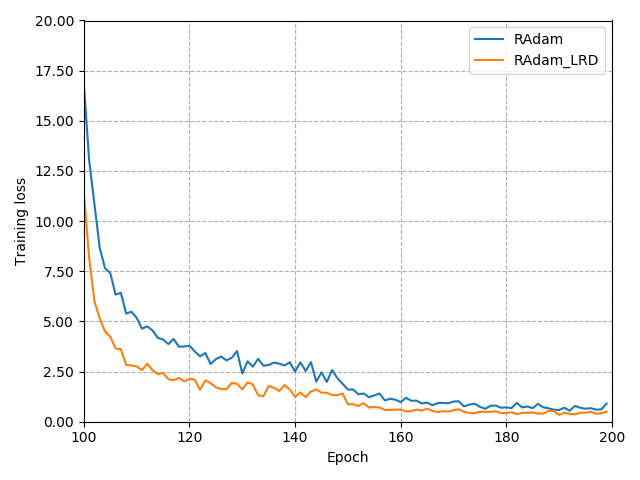}\\  
	\includegraphics[width=.19\textwidth]{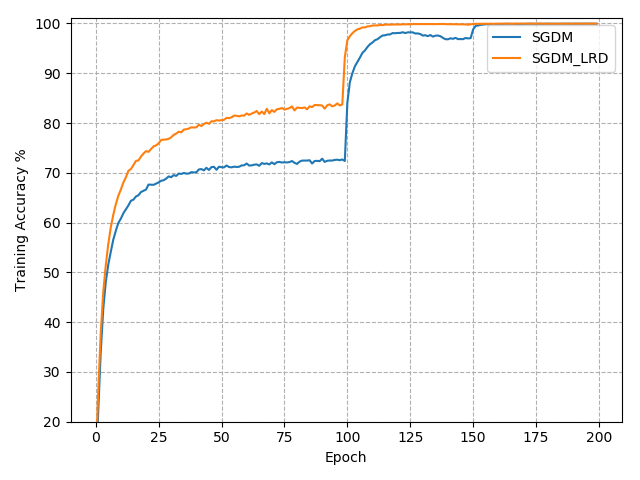}
	\includegraphics[width=.19\textwidth]{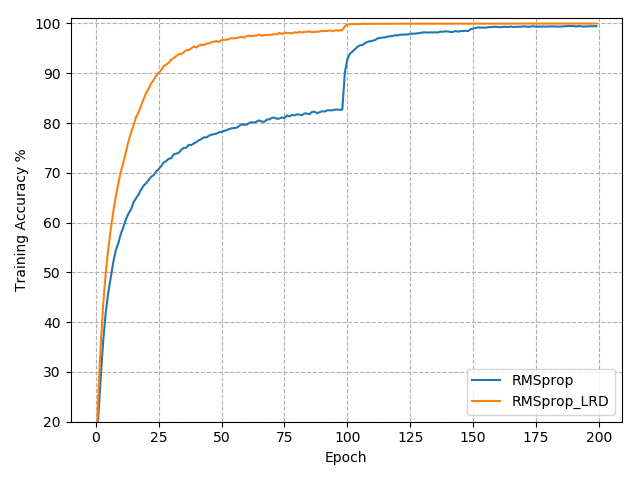}
	\includegraphics[width=.19\textwidth]{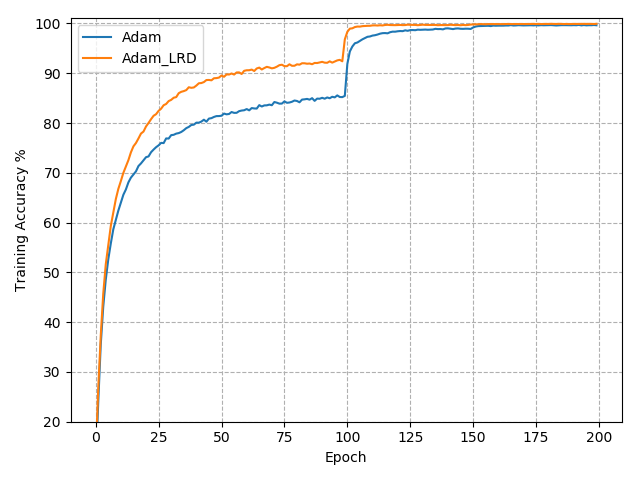}
	\includegraphics[width=.19\textwidth]{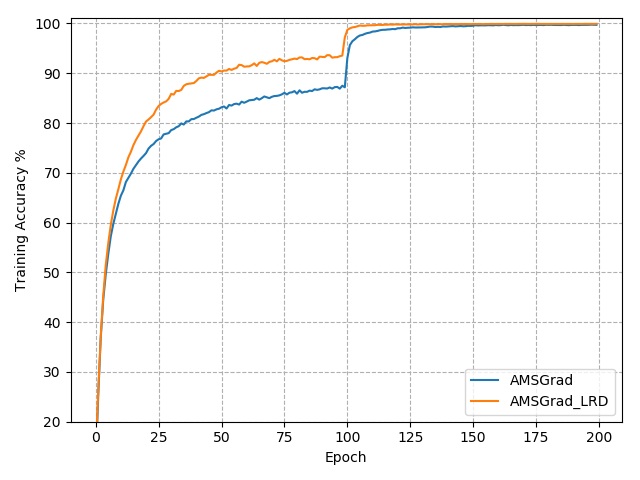}
	\includegraphics[width=.19\textwidth]{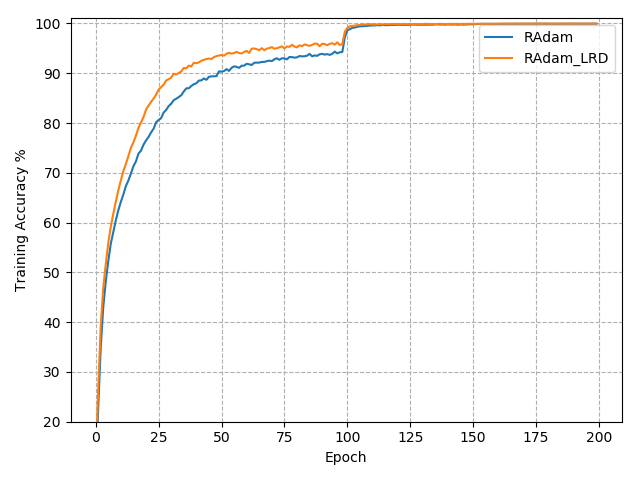}\\ 
	\includegraphics[width=.19\textwidth]{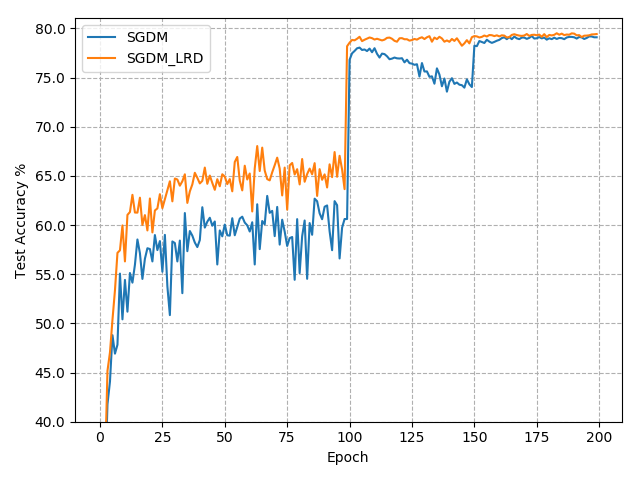}	
	\includegraphics[width=.19\textwidth]{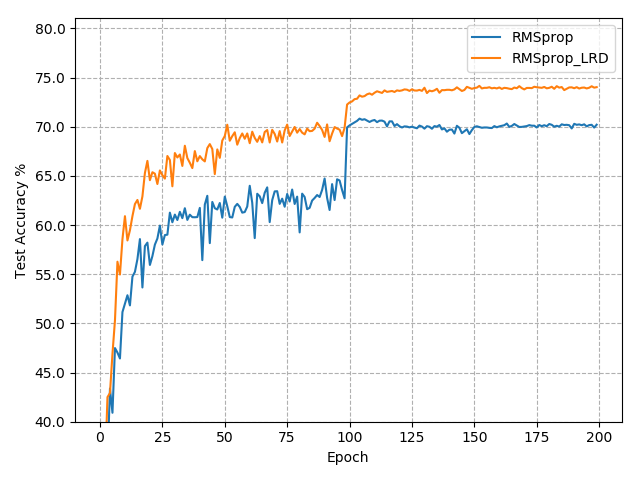}
	\includegraphics[width=.19\textwidth]{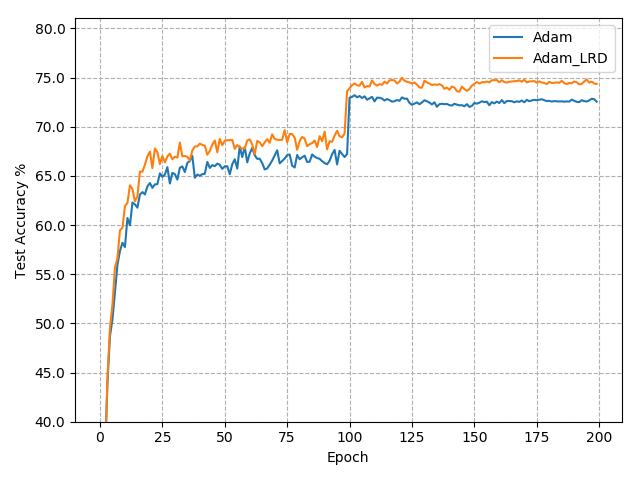}
	\includegraphics[width=.19\textwidth]{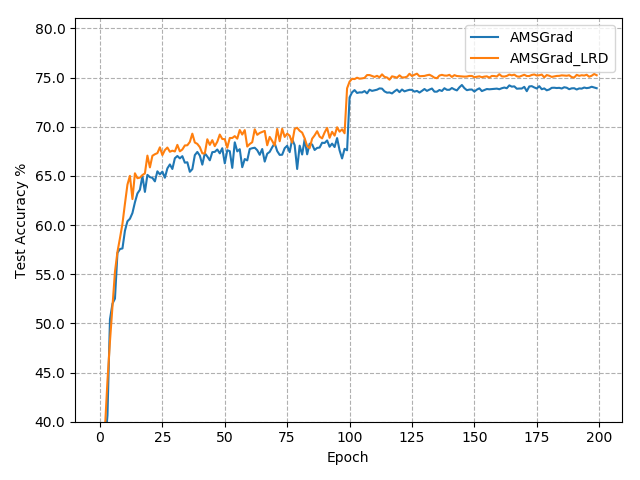}
	\includegraphics[width=.19\textwidth]{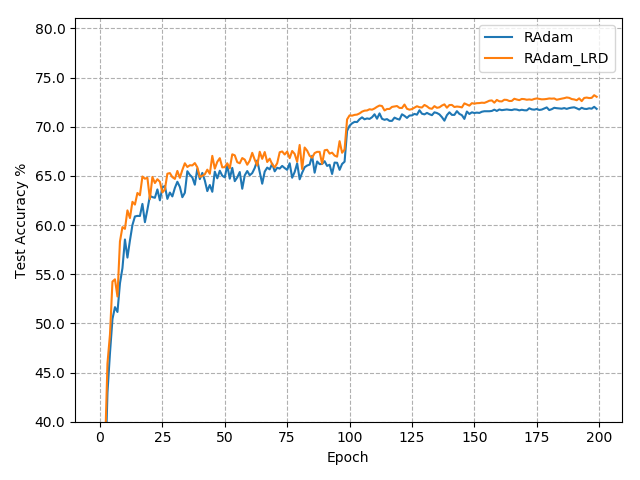}
	\caption{The learning curves for DenseNet-121 on CIFAR-100. Top: Training loss. Middle:  Training accuracy. Bottom: Test accuracy.}
	\label{cifar100}
\end{figure*}

\begin{table*}
	\caption{Test accuracy on image classification. ($\%$)}
	\centering
	
	\begin{tabular}{|c|c|c|c|c|c|c|}
		\hline
		\multirow{2}*{Dataset}& \multirow{2}*{Model} & \multicolumn{5}{|c|}{Optimization algorithms (without / with learning rate dropout)} \\
		\cline{3-7}
		& & SGDM $|$ LRD & RMSprop $|$ LRD &Adam $|$ LRD & AMSGrad $|$ LRD & RAdam $|$ LRD\\
		\hline
		MNIST&FCNet&97.82 / \textbf{98.21}&98.06 / \textbf{98.40}&97.88 / \textbf{98.15}&97.88 / \textbf{98.51}&97.97 / \textbf{98.21}\\
		\hline
		CIFAR-10&ResNet-34&95.30 / \textbf{95.54}&92.71 / \textbf{93.68}&93.05 / \textbf{93.77}&93.31 / \textbf{93.73}&94.24 / \textbf{94.61}\\
		\hline
		CIFAR-100&DenseNet-121&79.09 / \textbf{79.42}&70.21 / \textbf{74.02}&72.55 / \textbf{74.34}&73.91 / \textbf{75.23}&71.81 / \textbf{73.04}\\
		\hline
	\end{tabular}
	\label{table1}
\end{table*}

\subsection{Image classification}
We first apply learning rate dropout to the multiclass classification problem of the MNIST, CIFAR-10 and CIFAR-100 datasets.

\paragraph{MNIST:} The MNIST digits dataset \cite{lecun1998gradient} contains $60,000$ training and $10000$ test images of size $28\times 28$. The task is to classify the images into 10 digit classes. Following the previous publications \cite{Kingma2014Adam, an2018pid}, we train a simple 2-hidden fully connected layer neural network (FCNet) on MNIST. We use a fully connected 1000 rectified linear units (ReLU) as each hidden layer for this experiment. The training was on mini-batches with 128 images per batch for 100 epochs through the training set. A decay scheme is not used.

\paragraph{CIFAR:} The CIFAR-10 and CIFAR-100 datasets consist of 60,000 RGB images of size $32\times 32$, drawn from 10 and 100 categories, respectively. 50,000 images are used for training and the rest for testing. In both datasets, training and testing images are uniformly distributed over all the categories. To show the broad applicability of the proposed method, we use the ResNet-34 \cite{he2016deep} for CIFAR-10 and DenseNet-121 \cite{huang2017densely} for CIFAR-100. Both models are trained for 200 epoches with a mini-batch size 128. We reduce the learning rates by 10 times at the 100-th and 150-th epoches. The weight decay rate is set to $5e-4$.

To do a reliable evaluation, the classifiers are trained multiple times using several optimization algorithms including SGD-momentum (SGDM) \cite{qian1999momentum}, RMSprop \cite{tieleman2012lecture}, Adam \cite{Kingma2014Adam}, AMSGrad \cite{reddi2019convergence} and RAdam \cite{liu2019variance}. We apply learning rate dropout to each optimization algorithms to show how this technique can help with training. Unless otherwise specified, the dropout rate $p$ (the probability of performing an individual parameter update in any given iteration) is set to $0.5$. For each optimization algorithm, the initial learning rate has the greatest impact on the ultimate solution. Therefore, we only tune the initial learning rate while retaining the default settings for other hyper-parameters. Specifically, the initial learning rates for SGDM, RMSprop, Adam, AMSGrad, and RAdam are 0.1, 0.001, 0.001, 0.001, 0.03, respectively.

We first show the learning curves for MNIST in Figure \ref{mnist}. We find that all methods show fast convergence speed and good generalization. However, compared with their prototypes, the optimizers that use learning rate dropout still display better performance in terms of training speed and test accuracy. We further report the results for CIFAR datasets in Figure \ref{cifar10} and Figure \ref{cifar100}. The test accuracy obtained by different methods is summarized in Table \ref{table1}. The same models with and without learning rate dropout have very different convergence. These optimization algorithms without learning rate dropout either have slow training or converge to poor results. In contrast, learning rate dropout benefits all optimization algorithms by improving training speed and generalization. In particular we observe that by applying learning rate dropout to the non-adaptive method SGDM, we can achieve convergence speed comparable to other adaptive methods. On CIFAR-100, learning rate dropout even helps RMSprop obtain an accuracy improvement of close to $4\%$. Furthermore, learning rate dropout incurs negligible computational costs and no parameter tuning aside from the dropout rate $p$.


\subsection{Image segmentation}
\begin{figure}
	\centering
	\includegraphics[width=1.6in]{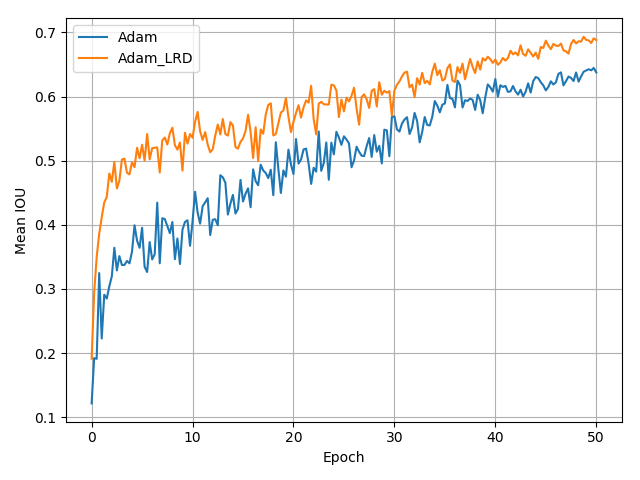}
	\includegraphics[width=1.6in]{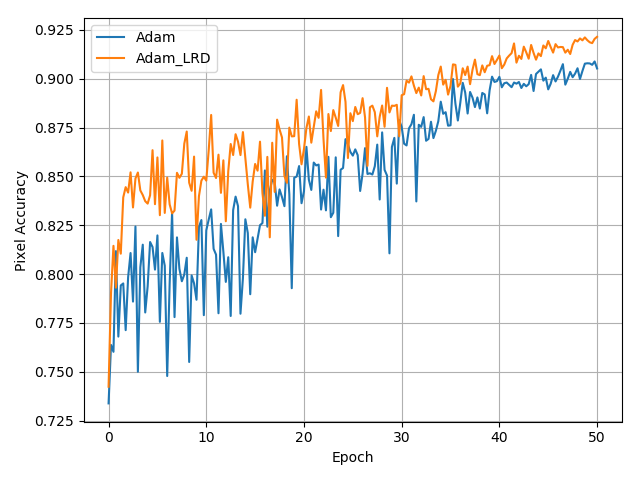}
	\caption{Results for PSPNet on VOC2012 semantic segmentation dataset. Left: mean IOU. Right: Pixel Accuracy.}
	\label{seg}
\end{figure}

We also consider the semantic segmentation task, which assigns each pixel in an image a semantic label \cite{chen2018encoder,long2015fully}. We evaluate our method on the PASCAL VOC2012 semantic segmentation dataset \cite{everingham2010pascal}, which consists of 20 object categories and one background category. Following the conventional setting in \cite{chen2014semantic, lin2016efficient}, the dataset is augmented by extra annotated VOC images provided in \cite{hariharan2011semantic}, which results in 10,582, 1,449 and 1,456 images for training, validation and testing, respectively. We use the state-of-the-art segmentation model PSPNet \cite{zhao2017pyramid} to conduct the experiments. We use Adam solver with the initial learning rate $0.001$. Other hyperparameters such as batch size and weight decay follow the setting in \cite{zhao2017pyramid}. The segmentation performance is 
measured by the mean of class-wise intersection over union (Mean IoU) and pixel-wise accuracy (Pixel Accuracy). Results for this experiment are reported in Figure \ref{seg}. With our proposed learning rate dropout, the model yields results 0.688/0.921 in terms of Mean IoU and Pixel Accuracy, exceeding the vanilla Adam of 0.637/0.905. In addition, Figure \ref{seg} shows once again that learning rate dropout leads to faster convergence.

\begin{figure}
	\centering
	\includegraphics[width=1.6in]{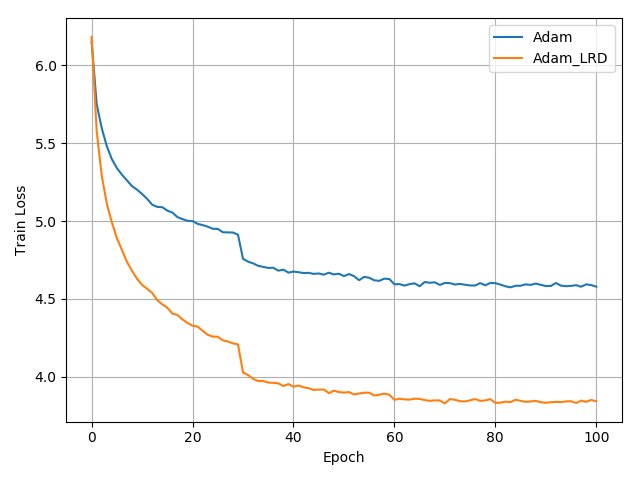}
	\includegraphics[width=1.6in]{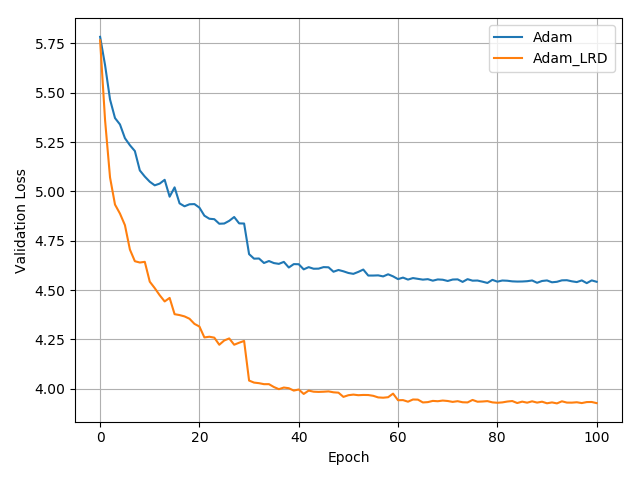}
	\caption{Results for object detection. Left: Training loss. Right: Validation loss.}
	\label{detect}
\end{figure}

\subsection{Object detection}
Object detection is a challenging and important problem in the computer vision community. We next apply learning rate dropout to an object detection task. We use VOC2012 $trainval$ and VOC2007 $trainval$ datasets for training, and use VOC2007 $test$ as our validation and testing data. We train a one-stage detection network SSD \cite{liu2016ssd} using Adam with a learning rate 0.001. Other hyperparameter settings are the same as in \cite{liu2016ssd}. To show the effect of learning rate dropout on training, we use the training loss and validation loss as the evaluation metric. We run the model for 100 epochs and show the results in Figure \ref{detect}. After applying learning rate dropout, the training loss and validation loss drop faster and converge to better results. This means that the detector achieves higher detection accuracy.

\begin{figure}
	\centering
	\includegraphics[width=1.6in]{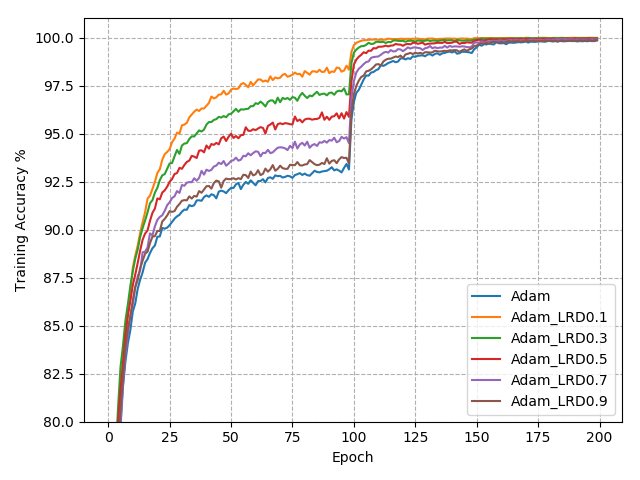}
	\includegraphics[width=1.6in]{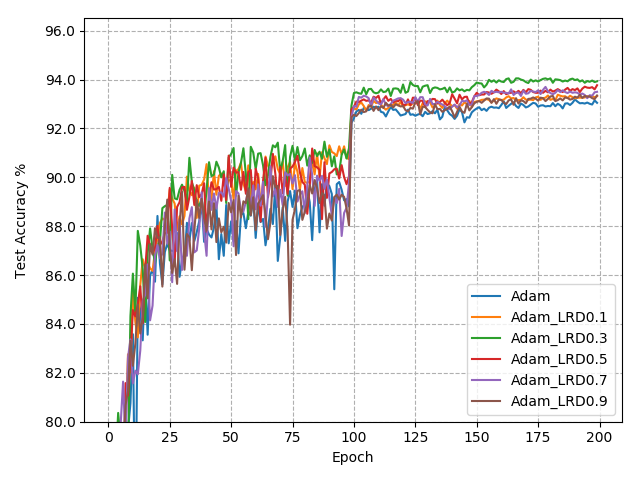}\\
	\subfigure[Training accuracy]{\includegraphics[width=1.6in]{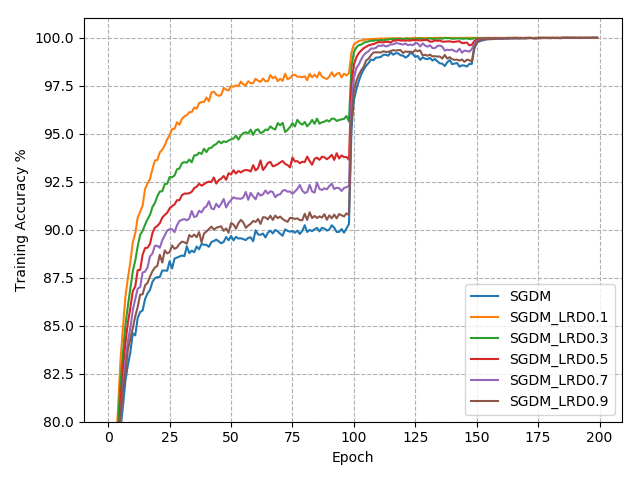}}
	\subfigure[Test accuracy]{\includegraphics[width=1.6in]{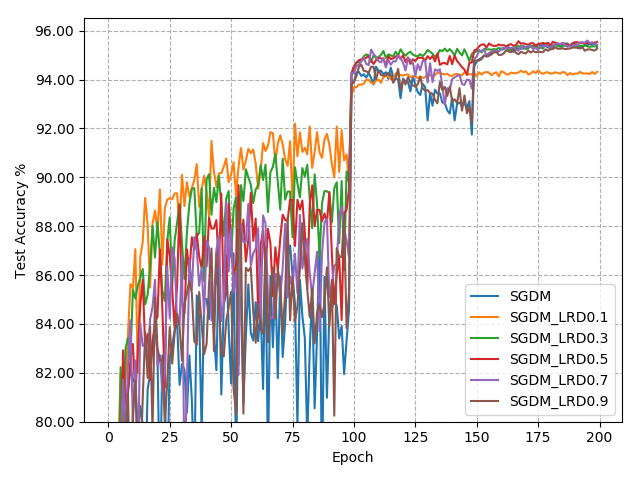}}
	\caption{Results obtained using different dropout rate $p$. Top: Adam. Bottom: SGDM.}
	\label{dropout_rate}
\end{figure}

\begin{table}
	\caption{Test accuracy on CIFAR-10 using different dropout rate $p$. ($\%$)}
	\centering
	
	\begin{tabular}{|c|c|c|}
		\hline
		{}& Adam &SGDM \\
		\hline
		$p=1$(No LRD) & 93.05 &95.30\\
		\hline
		$p=0.1$ &93.32 &94.32\\
		\hline
		$p=0.3$  &\textbf{93.93}&95.39\\
		\hline
		$p=0.5$  &93.77&\textbf{95.54}\\
		\hline
		$p=0.7$ & 93.50&95.42\\
		\hline
		$p=0.9$ &93.35&95.26\\
		\hline
	\end{tabular}
	\label{table_dropout_rate}
\end{table}

\subsection{Effect of dropout rate}
As mentioned, the hyperparameter $p$ which controls the expected size of the coordinate update set is the only parameter that needs to be tuned. In this section, we explore the effect of various $p$. We conduct experiments with ResNet-34 on the CIFAR-10 dataset, where $p$ is chosen from a broad range $\{0.1,0.3,0.5,0.7,0.9\}$. We use Adam and SGDM to train the model, other hyperparameters are the same as in previous experiments. The results are shown in Figure \ref{dropout_rate}. We can see that for any $p$, learning rate dropout can speed up training, but the smaller the $p$, the faster the convergence. In addition, we report the test accuracy in Table \ref{table_dropout_rate}. As can be seen, almost all values of $p$ lead to an improvement in test accuracy. In practical applications, in order to balance the training speed and generalization, we recommend setting $p$ in the range of $0.3$ to $0.7$.

\begin{figure}
	\centering
	\includegraphics[width=1.6in]{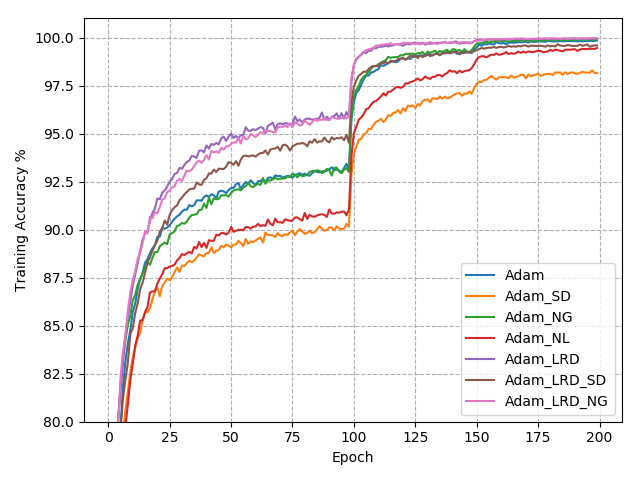}
	\includegraphics[width=1.6in]{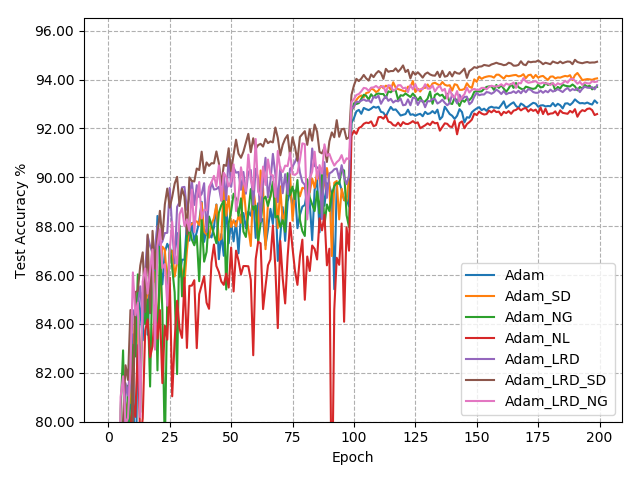}\\
	\subfigure[Training accuracy]{\includegraphics[width=1.6in]{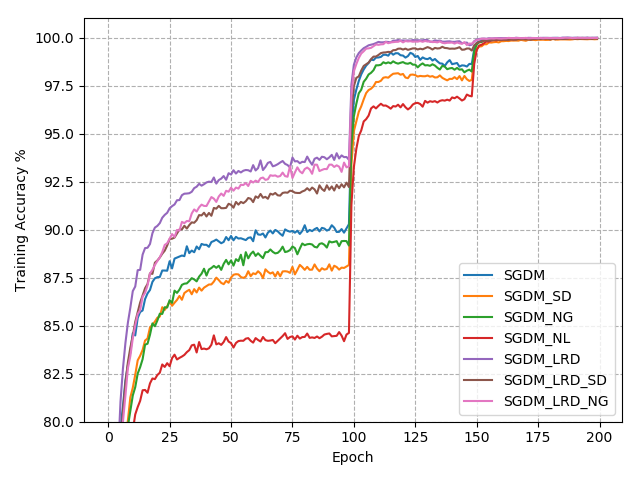}}
	\subfigure[Test accuracy]{\includegraphics[width=1.6in]{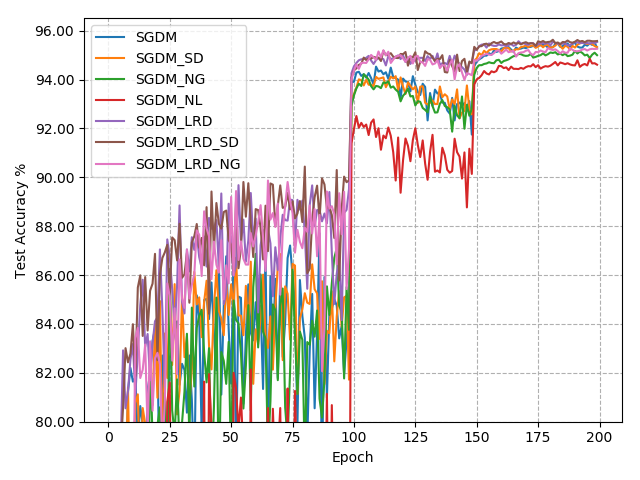}}
	\caption{Results on CIFAR-10 using different regularization strategies. Top: Adam. Bottom: SGDM.}
	\label{lrd_vs_sd}
\end{figure}

\subsection{Comparison with other regularizations}
We further compare our learning rate dropout with three popular regularization methods: standard dropout \cite{srivastava2014dropout}, noisy label \cite{xie2016disturblabel} and noisy gradient \cite{neelakantan2015adding}. The standard dropout regularizes the network on hidden units, and we set the probability that each hidden unit is retained to $0.9$. The noise label disturbs each train sample with the probability $0.05$ (\emph{i.e.}, a label is correct with a probability $0.95$). For each changed sample, the label is randomly drawn uniformly from the other labels except the correct one. The noisy gradient adds Gaussian noise $n \sim N(0,0.1)$ to the gradient at every time step. The variance of the Gaussian noise gradually decays to 0 according to the parameter setting in \cite{neelakantan2015adding}. We apply these regularization techniques to the ResNet-34 trained on CIFAR-10. This model is trained multiple times using SGDM and Adam. For clarity, we use the terms ``$\_SD$'', ``$\_NL$'', ``$\_NG$'' to denote 
training with standard dropout, noisy label and noisy gradient, respectively.

We report the learning curves in Figure \ref{lrd_vs_sd}. As can be seen, our learning rate dropout can speed up training while other regularization methods hinder convergence. This is because other regularization methods inject random noise into the training, which may lead to the wrong loss descent path. In contrast, our learning rate dropout always samples a correct loss descent path, so it does not worsen the training. We also show the test accuracy in Table \ref{table2}. The results show that the standard dropout and our learning rate dropout can effectively improve generalization, while the effects of noise label and noisy gradient are disappointing. On the other hand, we find that our learning rate dropout and other regularization methods can be complementary. We use ``$\_LRD\_**$'' to represent the simultaneous use of learning rate dropout and other optimizers. From Figure \ref{lrd_vs_sd} and Table \ref{table2}, we see that learning rate dropout can work with other regularization methods to further improve the performance of the model. This shows that learning rate dropout is not limited to being a substitute for standard dropout or other methods.

\begin{table}
	\caption{Test accuracy on CIFAR-10 using different regularization strategies. ($\%$)}
	\centering
	
	\begin{tabular}{|c|c|c|}
		\hline
		{}& Adam &SGDM \\
		\hline
		No regularization & 93.05 &95.30\\
		\hline
		Standard dropout (SD) &94.05 &95.33\\
		\hline
		Noise gradient (NG) & 93.70&95.00\\
		\hline
		Noise label (NL)&92.59&94.61\\
		\hline
		Learning rate dropout (LRD) &93.77&95.54\\
		\hline
		LRD and SD&\textbf{94.73}&\textbf{95.58}\\
		\hline
		LRD and NG&93.92&95.26\\
		\hline
	\end{tabular}
	\label{table2}
\end{table}

\begin{figure}
	\centering
	
	\includegraphics[width=1.6in]{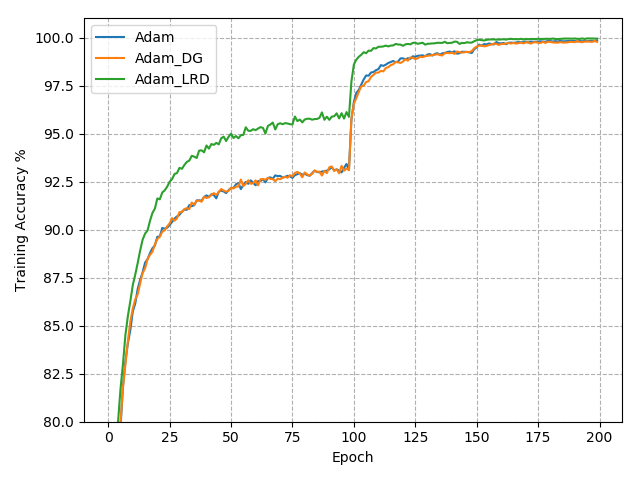}
	\includegraphics[width=1.6in]{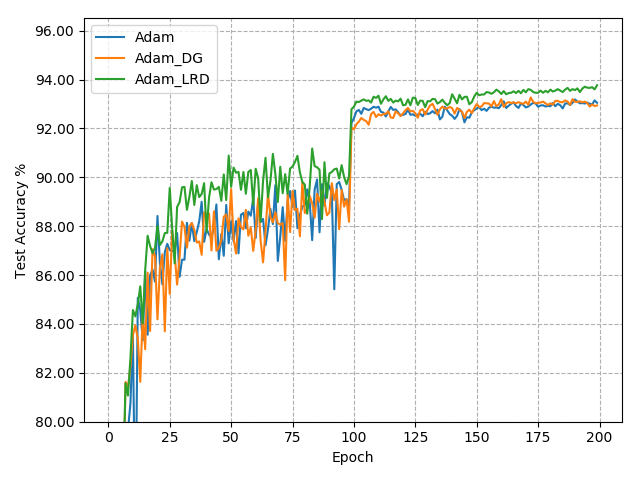}\\
	
	\subfigure[Training accuracy]{\includegraphics[width=1.6in]{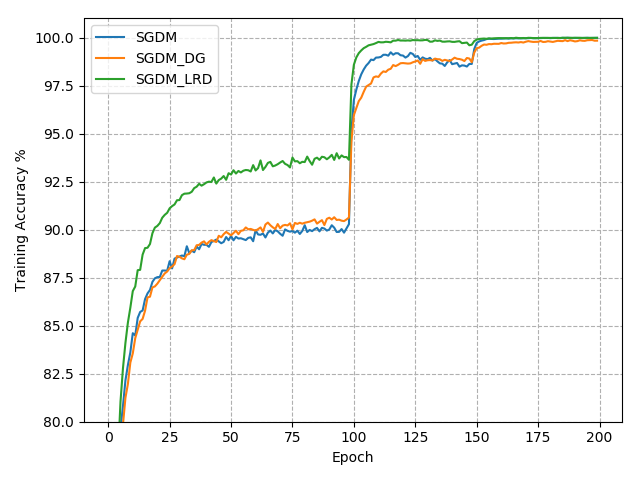}}
	\subfigure[Test accuracy]{\includegraphics[width=1.6in]{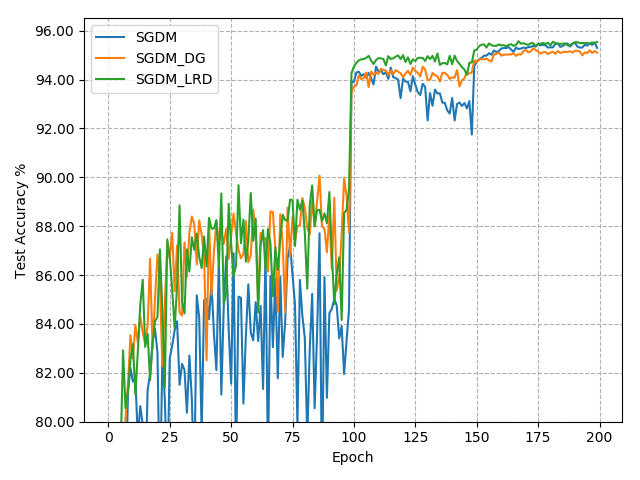}}
	\caption{LRD vs. DG on CIFAR-10 (ResNet-34 is used). Top: Adam. Bottom: SGDM.}
	\label{dg}
\end{figure}

\subsection{Dropout on $G_t$}
Learning rate dropout is also equivalent to applying dropout to the update $\triangle W_t$ of parameters $W$ at each timestep $t$. The idea is to inject uncertainty into the loss descent path, while ensuring that the loss descends in the correct direction at each iteration. One may wonder if there are other ways to realize this idea. For example, using dropout on the gradient $G_t=\triangledown f_t(W_{t-1})$ is also a possible solution. The dropout on $G_t$ can also interfere with the loss descent path and does not lead to the wrong descent direction. Here, we compare our learning rate dropout with dropout on $G_t$ (DG). DG follows the setting of LRD, the gradient $g_{ij,t}$ of a parameter $w_{ij,t-1}$ is retained with a probability $p$, or is set to 0 with probability $1-p$. In this experiments, $p=0.5$. We show the learning curves in Figure \ref{dg}. As can be seen , DG has no positive effect on training. The causes of this may be multiple. First, the gradient 
accumulation terms (\emph{e.g.}, momentum) can greatly suppress the impact of dropping gradients on the loss descent path. In addition, dropping the gradient may slow down training due to the lack of gradient information. In contrast, our LRD only temporarily stops updating some parameters, and all gradient information is stored by the gradient accumulation terms. Therefore, in our learning rate dropout training, there is no loss of gradient information.

\section{Dropout rate of standard dropout}
\begin{figure}[t]
	\centering
	
	\subfigure[Training accuracy]{\includegraphics[width=1.6in]{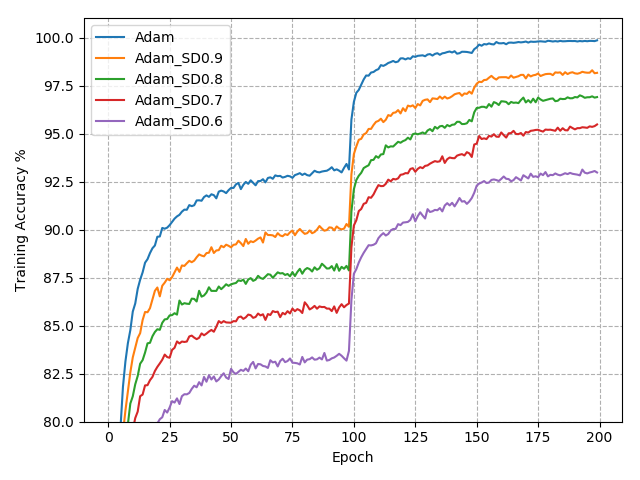}}
	\subfigure[ Test accuracy]{\includegraphics[width=1.6in]{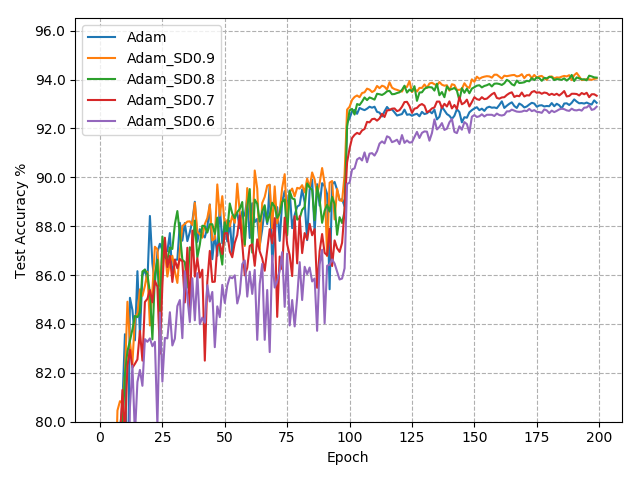}}
	\caption{Results obtained using different $p_{sd}$.}
	\label{psd_sup2}
\end{figure}

We have shown that LRD can speed up training and improve generalization with almost any dropout rate $p$ ($p\in (0,1)$). The standard dropout (SD) also has a tunable dropout rate $p_{sd}$. In this section, we explore the effect of various $p_{sd}$. We apply SD to ResNet-34 trained on CIFAR-10 dataset, where $p_{sd}$ is chosen from $\{0.9, 0.8, 0.7, 0.6\}$. We show the results in Figure \ref{psd_sup2} and Table \ref{table_psd2}. As can be seen, the performance of standard dropout is sensitive to the choice of dropout rate $p_{sd}$. The smaller the $p_{sd}$, the worse the convergence. This indicates that we should choose $p_{sd}$ carefully when applying SD to convolution neural networks.

\begin{table}
	\caption{Test accuracy on CIFAR-10 using different $p_{sd}$. ($\%$)}
	\centering
	
	\begin{tabular}{|c|c|c|c|c|c|}
		\hline
		$p_{sd}$& 1 (No SD) &0.9&0.8&0.7&0.6 \\
		\hline
		Adam & 93.05&94.05&\textbf{94.08}&93.34&92.89 \\
		\hline
		
	\end{tabular}
	\label{table_psd2}
\end{table}

\section{Conclusion}
We presented learning rate dropout, a new technique for regularizing neural network training with gradient descent. This technique encourages the optimizer to actively explore in the parameter space by randomly dropping some learning rates. The uncertainty of the learning rate helps models quickly escape poor local optima and gives models more opportunities to search for better results. Experiments show the substantial ability of learning rate dropout to accelerate training and enhance generalization. In addition, this technique is found to be effective in wide variety of application domains including image classification, image segmentation and object detection. This shows that learning rate dropout is a general technique, which has great potential in practical applications.

{\small
	\bibliographystyle{ieee}
	\bibliography{egbib}
}

\end{document}